\documentclass{article}

\usepackage{microtype}
\usepackage{graphicx}
\usepackage{amsfonts}
\usepackage{amssymb}
\usepackage{amsmath}
\usepackage{subfigure}
\usepackage{bm}
\usepackage{natbib} 
\usepackage{booktabs} 
\usepackage{multirow,array}

\DeclareMathOperator*{\argmax}{argmax}

\usepackage{xcolor}
\definecolor{myblue}{HTML}{0000B5}
\definecolor{crimson}{HTML}{B30000}
\newcommand{\bb}[1]{\textcolor{myblue}{#1}}
\newcommand{\cc}[1]{\textcolor{crimson}{#1}}

\usepackage{hyperref}

\usepackage[accepted]{include/icml2018}

\icmltitlerunning{QMIX: Monotonic Value Function Factorisation for
	Deep Multi-Agent Reinforcement Learning}

\begin{document}

\twocolumn[
\icmltitle{QMIX: Monotonic Value Function Factorisation for\\
	Deep Multi-Agent Reinforcement Learning}

\icmlsetsymbol{equal}{*}

\begin{icmlauthorlist}
\icmlauthor{Tabish Rashid}{equal,ox}
\icmlauthor{Mikayel Samvelyan}{equal,rau}
\icmlauthor{Christian Schroeder de Witt}{ox}\\
\icmlauthor{Gregory Farquhar}{ox}
\icmlauthor{Jakob Foerster}{ox}
\icmlauthor{Shimon Whiteson}{ox}
\end{icmlauthorlist}

\icmlaffiliation{ox}{University of Oxford, Oxford, United Kingdom}
\icmlaffiliation{rau}{Russian-Armenian University, Yerevan, Armenia}

\icmlcorrespondingauthor{Tabish Rashid}{tabish.rashid@cs.ox.ac.uk}
\icmlcorrespondingauthor{Mikayel Samvelyan}{mikayel@samvelyan.com}

\icmlkeywords{Machine Learning, ICML}

\vskip 0.3in
]

\printAffiliationsAndNotice{\icmlEqualContribution}

\begin{abstract}
\label{sec:abstract}

In many real-world settings, a team of agents must coordinate their behaviour  
while acting in a decentralised way. At the same time, it is often possible to 
train the agents in a centralised fashion in a simulated or laboratory setting, 
where global state information is available and communication constraints are lifted. 
Learning joint action-values conditioned on extra state information is 
an attractive way to exploit centralised learning, but the best strategy for 
then extracting decentralised policies is unclear.
Our solution is QMIX, a novel value-based method that can train decentralised policies in a centralised end-to-end fashion. QMIX employs a network that estimates joint action-values as a complex non-linear combination of per-agent values that condition only on local observations. 
We structurally enforce that the joint-action value is monotonic in the 
per-agent values, which allows tractable maximisation of the joint action-value in off-policy learning, and guarantees consistency between the 
centralised and decentralised policies.
We evaluate QMIX on a challenging set of StarCraft II micromanagement tasks, 
and show that QMIX significantly outperforms existing 
value-based multi-agent reinforcement learning methods.

\end{abstract} \section{Introduction}
\label{sec:intro}

Reinforcement learning (RL) holds considerable promise to help address a variety of cooperative multi-agent problems, such as coordination of robot swarms \cite{huttenrauch_guided_2017} and autonomous cars \cite{cao_overview_2012}. 

In many such settings, partial observability and/or communication constraints necessitate the learning of \textit{decentralised policies}, which condition only on the local action-observation history of each agent. Decentralised policies also naturally attenuate the problem that joint action spaces grow exponentially with the number of agents, often rendering the application of traditional single-agent RL methods impractical.

Fortunately, decentralised policies can often be learned in a centralised fashion in a simulated or laboratory setting. This often grants access to additional state information, otherwise hidden from agents, and removes inter-agent communication constraints. 
The paradigm of \textit{centralised training with decentralised execution} \cite{oliehoek_optimal_2008,kraemer_multi-agent_2016} has recently attracted attention in the RL community \cite{jorge_learning_2016,foerster_counterfactual_2017}. 
However, many challenges surrounding how to best exploit centralised training remain open.

One of these challenges is how to represent and use the action-value function that most RL methods learn.  On the one hand, properly capturing the effects of the agents' actions requires a centralised action-value function $Q_{tot}$ that conditions on the global state and the joint action.  On the other hand, such a function is difficult to learn when there are many agents and, even if it can be learned, offers no obvious way to extract decentralised policies that allow each agent to select only an individual action based on an individual observation.

\begin{figure}[t!]
	\centering
		\subfigure[5 Marines map]{
		\includegraphics[width=0.4842\columnwidth]{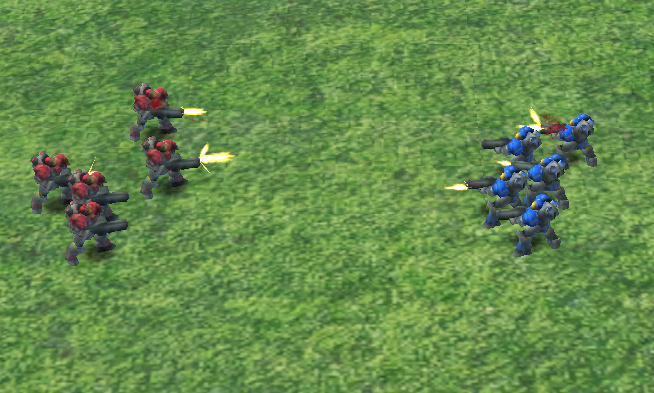}}
			\subfigure[2 Stalkers \& 3 Zealots map]{
	\includegraphics[width=0.4842\columnwidth]{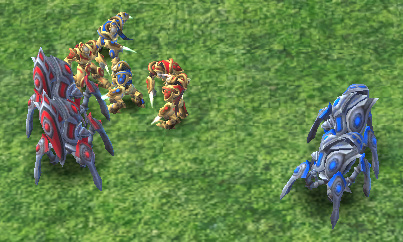}}
	\caption{\textit{Decentralised unit micromanagement} in StarCraft II, where each learning agent controls an individual unit. The goal is to coordinate behaviour across agents to defeat all enemy units.}
	\label{fig:starcraft_screenshots}
\end{figure}

The simplest option is to forgo a centralised action-value function and let each agent $a$ learn an individual action-value function $Q_a$ independently, as in \emph{independent Q-learning} (IQL) \cite{tan_multi-agent_1993}.  However, this approach cannot explicitly represent interactions between the agents and may not converge, as each agent's learning is confounded by the learning and exploration of others.

At the other extreme, we can learn a fully centralised state-action value function $Q_{tot}$ and then use it to guide the optimisation of decentralised policies in an actor-critic framework, an approach taken by \emph{counterfactual multi-agent} (COMA) policy gradients \cite{foerster_counterfactual_2017}, as well as work by \citet{gupta_cooperative_2017}. However, this requires on-policy learning, which can be sample-inefficient, and training the fully centralised critic becomes impractical when there are more than a handful of agents.

In between these two extremes, we can learn a centralised but factored $Q_{tot}$, an approach taken by \emph{value decomposition networks} (VDN) \cite{sunehag_value-decomposition_2017}. By representing $Q_{tot}$ as a sum of individual value functions $Q_a$ that condition only on individual observations and actions, a decentralised policy arises simply from each agent selecting actions greedily with respect to its $Q_a$. However, VDN severely limits the complexity of centralised action-value functions that can be represented and ignores any extra state information available during training.

In this paper, we propose a new approach called QMIX which, like VDN, lies between the extremes of IQL and COMA, but can represent a much richer class of action-value functions. Key to our method is the insight that the full factorisation of VDN is not necessary to extract decentralised policies.  Instead, we only need to ensure that a global $\argmax$ performed on $Q_{tot}$ yields the same result as a set of individual $\argmax$ operations performed on each $Q_a$.  To this end, it suffices to enforce a monotonicity constraint on the relationship between $Q_{tot}$ and each $Q_a$:
\begin{equation}
\label{eq:monotonicity_constraint}
\frac{\partial Q_{tot}}{\partial Q_a}  \geq 0,~ \forall a.
\end{equation}

QMIX consists of \textit{agent networks} representing each $Q_a$, and a \emph{mixing network} that combines them into $Q_{tot}$, not as a simple sum as in VDN, but in a complex non-linear way that ensures consistency between the centralised and decentralised policies. At the same time, it enforces the constraint of \eqref{eq:monotonicity_constraint} by restricting the mixing network to have positive weights.  
As a result, QMIX can represent complex centralised action-value functions with a factored representation that scales well in the number of agents and allows decentralised policies to be easily extracted via linear-time individual argmax operations.

We evaluate QMIX on a range of unit micromanagement tasks built in StarCraft II\footnote{StarCraft and StarCraft II are trademarks of Blizzard Entertainment\textsuperscript{TM}.}. \cite{vinyals_starcraft_2017}. Our experiments show that QMIX outperforms IQL and VDN, both in terms of absolute performance and learning speed.
In particular, our method shows considerable performance gains on a task with heterogeneous agents. Moreover, our ablations show both the necessity of conditioning on the state information and the non-linear mixing of agent $Q$-values in order to achieve consistent performance across tasks.

 \section{Related Work}
\label{sec:related}

Recent work in multi-agent RL has started moving from 
tabular methods \cite{yang_multiagent_2004, busoniu_comprehensive_2008} to deep learning methods that can tackle
high-dimensional state and action spaces \cite{tampuu_multiagent_2015,foerster_counterfactual_2017,peng_multiagent_2017}. In this paper, we 
focus on cooperative settings.

On the one hand, a natural approach to finding policies for a multi-agent system 
is to directly learn decentralised value functions or policies. 
\emph{Independent Q-learning} \cite{tan_multi-agent_1993} trains independent
action-value functions for each agent using $Q$-learning \cite{watkins_learning_1989}. \cite{tampuu_multiagent_2015} extend this approach to 
deep neural networks using DQN \cite{mnih_human-level_2015}.
While trivially achieving decentralisation, these approaches are prone to instability arising from the non-stationarity of the environment 
induced by simultaneously learning and exploring agents. 
\citet{omidshafiei_deep_2017} and \citet{foerster_stabilising_2017} address 
learning stabilisation to some extent, but still learn decentralised value 
functions and do not allow for the inclusion of extra state information during training.

On the other hand, centralised learning of joint actions can naturally handle 
coordination problems and avoids non-stationarity, but is hard to scale, as the joint action space grows exponentially in the 
number of agents.
Classical approaches to scalable centralised learning include 
\textit{coordination graphs} \cite{guestrin_multiagent_2002}, which exploit conditional independencies between agents by decomposing a global reward function 
into a sum of agent-local terms.
\textit{Sparse cooperative Q-learning} \cite{kok_collaborative_2006} is a tabular $Q$-learning 
algorithm that learns to coordinate the actions of a group of cooperative 
agents only in the states in which such coordination is necessary, encoding
those dependencies in a coordination graph. 
These methods require the dependencies between agents to be pre-supplied, whereas we do not require such prior knowledge. 
Instead, we assume that each agent always contributes to the global reward, and learns the magnitude of its contribution in each state.

More recent approaches for centralised learning require even more 
communication during execution: CommNet 
\cite{sukhbaatar_learning_2016} uses a centralised network architecture to exchange information between agents. BicNet 
\cite{peng_multiagent_2017} uses bidirectional RNNs to exchange 
information between agents in an actor-critic setting. This approach 
additionally requires estimating individual agent rewards.

Some work has developed hybrid approaches that exploit the setting of 
centralised learning with fully decentralised execution. COMA 
\cite{foerster_counterfactual_2017} uses a centralised critic to train 
decentralised actors, estimating a counterfactual advantage function for each 
agent in order to address multi-agent credit assignment. Similarly, 
\citet{gupta_cooperative_2017} present a centralised actor-critic algorithm 
with per-agent critics, which scales better with the number of agents but mitigates the
advantages of centralisation. \citet{lowe_multi-agent_2017} learn a 
centralised critic for each agent and apply this to competitive games with 
continuous action spaces. These approaches use on-policy policy gradient 
learning, which can have poor sample efficiency and is prone to getting stuck 
in sub-optimal local minima.

\citet{sunehag_value-decomposition_2017} propose \emph{value decomposition 
networks} (VDN), which allow for centralised value-function learning with 
decentralised execution. Their algorithm decomposes a central 
state-action value function into a sum of individual agent terms. This corresponds to 
the use of a degenerate fully disconnected coordination graph. VDN does not make 
use of additional state information during training and can represent only a
 limited class of centralised action-value functions.
 
A number of papers have established unit micromanagement in StarCraft as a benchmark for deep multi-agent RL. 
\citet{usunier_episodic_2016} present an algorithm using a centralised \textit{greedy MDP} and first-order optimisation. \citet{peng_multiagent_2017} also evaluate their methods on StarCraft. However, neither requires decentralised execution. 
Similar to our setup is the work of \citet{foerster_stabilising_2017}, who evaluate replay stabilisation methods for IQL on combat scenarios with up to five agents. \citet{foerster_counterfactual_2017} also uses this setting. 
In this paper, we construct unit micromanagement tasks in the \textit{StarCraft 
II Learning Environment} (SC2LE) \cite{vinyals_starcraft_2017} as opposed to 
StarCraft, because it is actively supported by the game developers and SC2LE offers a more stable testing environment. 

QMIX relies on a neural network to transform the centralised state into the 
weights of another neural network, in a manner reminiscent of 
\emph{hypernetworks} \citep{ha_hypernetworks_2016}. This second neural network 
is constrained to be monotonic with respect to its inputs by keeping its 
weights positive. \citet{Dugas_2009} investigate such functional restrictions 
for neural networks.

 \section{Background}
\label{sec:background}

A \textit{fully cooperative multi-agent task} can be described as a Dec-POMDP \cite{oliehoek_concise_2016} consisting of a tuple $G=\left\langle S,U,P,r,Z,O,n,\gamma\right\rangle$. 
$s \in S$ describes the true state of the environment.
At each time step, each agent $a \in A \equiv \{1,...,n\}$ chooses an action $u^a\in U$, forming a joint action $\mathbf{u}\in\mathbf{U}\equiv U^n$. 
This causes a transition on the environment according to the state transition function $P(s'|s,\mathbf{u}):S\times\mathbf{U}\times S\rightarrow [0,1]$. 
All agents share the same reward function $r(s,\mathbf{u}):S\times\mathbf{U}\rightarrow\mathbb{R}$ and $\gamma\in[0,1)$ is a discount factor. 

We consider a \textit{partially observable} scenario in which each agent draws individual observations $z\in Z$ according to observation function $O(s,a):S\times A\rightarrow Z$. 
Each agent has an action-observation history $\tau^a\in T\equiv(Z\times U)^*$, on which it conditions a stochastic policy $\pi^a(u^a|\tau^a):T\times U\rightarrow [0,1]$. The joint policy $\pi$ has a joint \textit{action-value function}: $Q^\pi(s_t, \mathbf{u}_t)=\mathbb{E}_{s_{t+1:\infty},\mathbf{u}_{t+1:\infty}} \left[R_t|s_t,\mathbf{u}_t\right]$, where $R_t=\sum^{\infty}_{i=0}\gamma^ir_{t+i}$ is the \textit{discounted return}.

Although training is centralised, execution is decentralised, i.e., the 
learning algorithm has access to all local action-observation histories 
$\boldsymbol{\tau}$ and global state $s$, but each agent's learnt policy can 
condition only on its own action-observation history $\tau^a$.

\subsection{Deep $Q$-Learning}

Deep $Q$-learning represents the action-value function with a deep neural network parameterised by $\theta$. \textit{Deep Q-networks} (DQNs) \cite{mnih_human-level_2015} use a  \textit{replay memory} to store the transition tuple $\left\langle s,u,r,s'\right\rangle$, where the state  $s'$  is observed after taking the action $u$ in state $s$ and receiving reward $r$. $\theta$ is learnt by sampling batches of $b$ transitions from the replay memory and minimising the squared \textit{TD error}:
\begin{equation}\label{eq:dqn}
\mathcal{L}(\theta)=\sum\limits_{i=1}^{b}\left[\left(y_i^{\text{DQN}}-Q(s,u;\theta)\right)^2\right],
\end{equation} 
where $y^{\text{DQN}}=r+\gamma\max_{u'} Q(s',u';\theta^-)$. $\theta^-$ are the parameters of a \textit{target network} that are periodically copied from $\theta$ and kept constant for a number of iterations.  

\subsection{Deep Recurrent $Q$-Learning}

In partially observable settings, agents can benefit from conditioning on their entire action-observation history. \citet{hausknecht_deep_2015} propose \textit{Deep Recurrent Q-networks} (DRQN) that make use of recurrent neural networks. Typically, gated architectures such as LSTM \cite{hochreiter_long_1997} or GRU \cite{chung_empirical_2014} are used to facilitate learning over longer timescales.

\subsection{Independent $Q$-Learning}

Perhaps the most commonly applied method in multi-agent learning is \textit{independent Q-learning} (IQL) \cite{tan_multi-agent_1993}, which decomposes a multi-agent problem into a collection of simultaneous single-agent problems that share the same environment. This approach does not address the non-stationarity introduced due to the changing policies of the learning agents, and thus, unlike $Q$-learning, has no convergence guarantees even in the limit of infinite exploration. In practice, nevertheless, IQL commonly serves as a surprisingly strong benchmark even in mixed and competitive games \cite{tampuu_multiagent_2015, leibo_multi-agent_2017}.

\subsection{Value Decomposition Networks}

By contrast, \textit{value decomposition networks} (VDNs) \citep{sunehag_value-decomposition_2017} aim to learn a joint action-value function $Q_{tot}(\boldsymbol{\tau},\mathbf{u})$, where $ \bm{\tau} \in \mathbf{T} \equiv \mathcal{T}^n $ is a joint action-observation history and $ \mathbf{u} $ is a joint action.  It represents $Q_{tot}$ as a sum of individual value functions $Q_a (\tau^a, u^a;\theta^a)$, one for each agent $a$, that condition only on individual action-observation histories:
\begin{equation}\label{eq:vdn}
Q_{tot}(\boldsymbol{\tau}, \mathbf{u}) = \sum_{i=1}^n Q_i (\tau^i, u^i;\theta^i).
\end{equation}
Strictly speaking, each $Q_a$ is a \textit{utility function} \cite{guestrin_multiagent_2002} and not a value function since by itself it does not estimate an expected return.  However, for terminological simplicity we refer to both $Q_{tot}$ and $Q_a$ as value functions.

The loss function for VDN is equivalent to \eqref{eq:dqn}, where $Q$ is replaced by $Q_{tot}$.  An advantage of this representation is that a decentralised policy arises simply from each agent performing greedy action selection with respect to its $Q_a$. 
 \section{QMIX}

\begin{figure*}[h!tb]
	\centering
	\includegraphics[width=\textwidth]{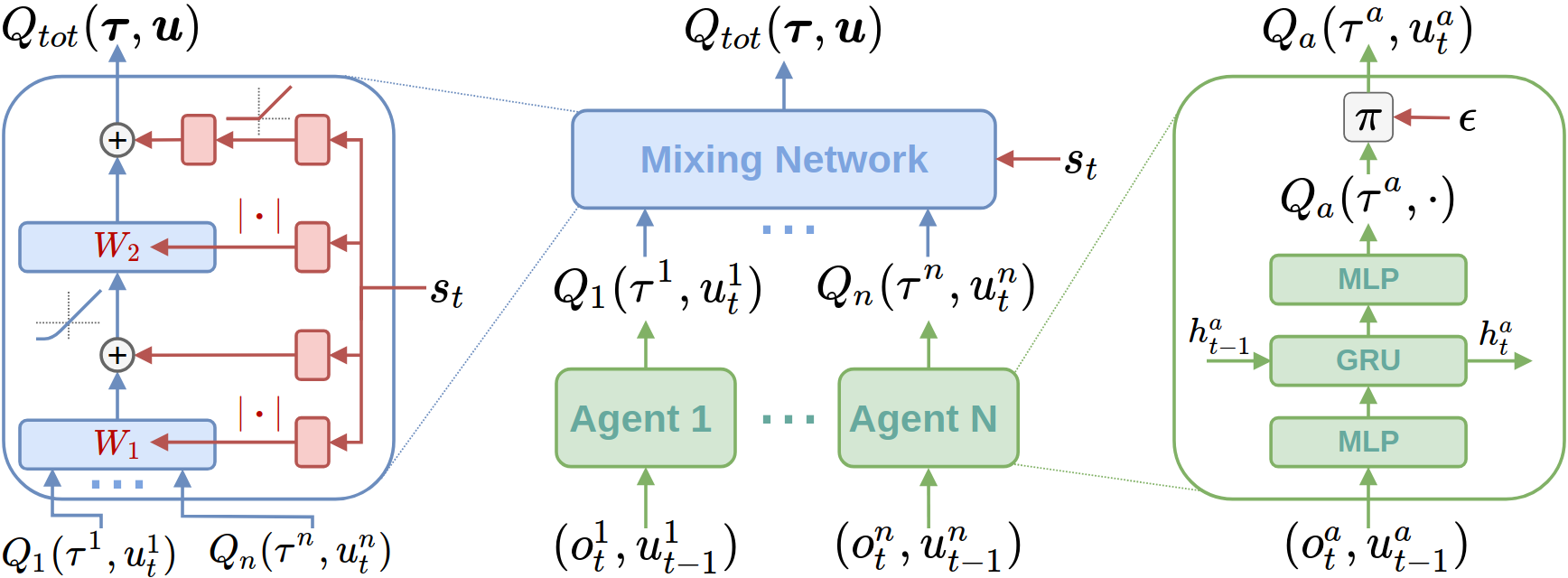}
	\text{~~~~~~~~~~~~~~~~~~~~~~~}(a) \hfill (b) \hfill (c) \text{~~~~~~~~~~~~~~~~~~~~}
	\caption{(a) Mixing network structure. In red are the hypernetworks that produce the weights and biases for mixing network layers shown in blue. (b) The overall QMIX architecture. (c) Agent network structure. Best viewed in colour.}
	\label{fig:QMIX}
\end{figure*}

\label{sec:methods}

In this section, we propose a new approach called QMIX which, like VDN, lies 
between the extremes of IQL and centralised $Q$-learning, but can represent a much richer class of action-value functions. 

Key to our method is the insight that the full factorisation of VDN is not necessary in order to be able to extract decentralised policies that are fully consistent with their centralised counterpart. Instead, for consistency we only need to ensure that a global $\argmax$ performed on $Q_{tot}$ yields the same result as a set of individual $\argmax$ operations performed on each $Q_a$:
\begin{equation}
\label{eq:argmax_constist}
\argmax_{\mathbf{u}}Q_{tot}(\boldsymbol{\tau}, \mathbf{u}) = 
\begin{pmatrix}
\argmax_{u^1}Q_1(\tau^1, u^1)   \\
\vdots \\
\argmax_{u^n}Q_n(\tau^n, u^n) \\
\end{pmatrix}.
\end{equation}
This allows each agent $a$ to participate in a decentralised execution solely by choosing greedy actions with respect to its $ Q_a $. As a side effect, if \eqref{eq:argmax_constist} is satisfied, then taking the $\argmax$ of $Q_{tot}$, required by off-policy learning updates, is trivially tractable. 

VDN's representation is sufficient to satisfy \eqref{eq:argmax_constist}. However, QMIX is based on the observation that this representation can be generalised to the larger family of monotonic functions that are also sufficient but not necessary to satisfy \eqref{eq:argmax_constist}.   
Monotonicity can be enforced through a constraint on the relationship between $Q_{tot}$ and each $Q_a$:
\begin{equation}
\label{eq:deriv-constr}
\frac{\partial Q_{tot}}{\partial Q_a}  \geq 0,~ \forall a \in A.
\end{equation}
To enforce \eqref{eq:deriv-constr}, QMIX represents $Q_{tot}$ using an architecture consisting of \textit{agent networks}, a \textit{mixing network}, and a set of \emph{hypernetworks} \cite{ha_hypernetworks_2016}. Figure \ref{fig:QMIX} illustrates the overall setup. 

For each agent $a$, there is one agent network that represents its individual value function $Q_a (\tau^a, u^a)$. We represent agent networks as DRQNs that receive the current individual observation $o^a_t$ and the last action $u^a_{t-1}$ as input at each time step, as shown in Figure \ref{fig:QMIX}c. 

The mixing network is a feed-forward neural network that takes the agent network outputs as input and mixes them monotonically, producing the values of $Q_{tot}$, as shown in Figure \ref{fig:QMIX}a. To enforce the monotonicity constraint of \eqref{eq:deriv-constr}, the weights (but not the biases) of the mixing network are restricted to be non-negative. This allows the mixing network to approximate any monotonic function arbitrarily closely \citep{Dugas_2009}.

The weights of the mixing network are produced by separate hypernetworks. Each hypernetwork takes the state $s$ as input and generates the weights of one layer of the mixing network. Each hypernetwork consists of a single linear layer, followed by an absolute activation function, to ensure that the mixing network weights are non-negative. The output of the hypernetwork is then a vector, which is reshaped into a matrix of appropriate size. The biases are produced in the same manner but are not restricted to being non-negative. The final bias is produced by a 2 layer hypernetwork with a ReLU non-linearity. Figure \ref{fig:QMIX}a illustrates the mixing network and the hypernetworks.

The state is used by the hypernetworks rather than being passed directly into the mixing network because $Q_{tot}$ is allowed to depend on the extra state 
information in non-monotonic ways. Thus, it would be overly constraining to pass some function of $s$ through the monotonic network alongside the 
per-agent values.
Instead, the use of hypernetworks makes it possible to condition the 
weights of the monotonic network on $s$ in an arbitrary way, thus 
integrating the full state $s$ into the joint action-value estimates as 
flexibly as possible.

QMIX is trained end-to-end to minimise the following loss:
\begin{equation}\label{eq:qmix_loss}
\mathcal{L}(\theta)=\sum\limits_{i=1}^b\left[\left(y_i^{tot} - Q_{tot}(\boldsymbol{\tau}, \mathbf{u}, s; \theta) \right)^2\right],
\end{equation} 
where $b$ is the batch size of transitions sampled from the replay buffer, $y^{tot} = r+\gamma\max_{\mathbf{u}'} Q_{tot}(\boldsymbol{\tau}', \mathbf{u}', s'; \theta^-)$ and $\theta^-$ are the parameters of a target network as in DQN. 
\eqref{eq:qmix_loss} is analogous to the standard DQN loss of \eqref{eq:dqn}. Since \eqref{eq:argmax_constist} holds, we can perform the maximisation of $Q_{tot}$ in time linear in the number of agents (as opposed to scaling exponentially in the worst case). 

\subsection{Representational Complexity}

The value function class representable with QMIX includes any value function that can be factored into a non-linear monotonic combination of the agents' individual value functions in the fully observable setting. 
This expands upon the linear monotonic value functions that are representable by VDN. 
However, the constraint in \eqref{eq:deriv-constr} prevents QMIX from representing value functions that do not factorise in such a manner.

Intuitively, any value function for which an agent's best action depends on the actions of the other agents \emph{at the same time step} will not factorise appropriately, and hence cannot be represented perfectly by QMIX. However, QMIX can approximate such value functions more accurately than VDN. Furthermore, it can take advantage of the extra state information available during training, which we show empirically. A more detailed discussion on the representation complexity is available in the supplementary materials.

 \section{Two-Step Game}
\label{sec:two_step_game}

To illustrate the effects of representational complexity of VDN and QMIX, we devise a simple two-step cooperative matrix game for two agents. 

At the first step, Agent $1$ chooses which of the two matrix games to play in the next timestep. For the first time step, the actions of Agent $2$ have no effect. In the second step, both agents choose an action and receive a global reward according to the payoff matrices depicted in Table \ref{tab:2step_game}.

We train VDN and QMIX on this task for $5000$ episodes and examine the final learned value functions in the limit of full exploration ($\epsilon=1$). Full exploration ensures that each method is guaranteed to eventually explore all available game states, such that the representational capacity of the state-action value function approximation remains the only limitation.
The full details of the architecture and hyperparameters used are provided in the supplementary material.

\begin{table}
	\setlength{\extrarowheight}{3pt}
	\centering
	\begin{tabular}{cc|*{2}{>{\centering\arraybackslash}p{.05\linewidth}|}}
		& \multicolumn{1}{c}{} & \multicolumn{2}{c}{\bb{Agent $2$}} \\
		& \multicolumn{1}{c}{} & \multicolumn{1}{c}{\bb{$A$}}  & \multicolumn{1}{c}{\bb{$B$}} \\ 
		\cline{3-4}
        \multirow{2}{*}{\rotatebox[origin=c]{90}{\cc{Agent $1$}}} & \cc{$A$} & 7 & 7 \\ \cline{3-4}
        & \cc{$B$} & 7 & 7  \\\cline{3-4}
        & \multicolumn{1}{c}{}  & \multicolumn{2}{c}{State $2$A} \\
    \end {tabular}~~~~~~~
    \begin{tabular}{cc|*{2}{>{\centering\arraybackslash}p{.05\linewidth}|}}
    	& \multicolumn{1}{c}{} & \multicolumn{2}{c}{\bb{Agent $2$}} \\
        & \multicolumn{1}{c}{} & \multicolumn{1}{c}{\bb{$A$}}  & \multicolumn{1}{c}{\bb{$B$}} \\ 
        \cline{3-4}
		\multirow{2}{*}{\rotatebox[origin=c]{90}{\cc{Agent $1$}}} & \cc{$A$} & 0 & 1 \\ \cline{3-4}
		& \cc{$B$} & 1 & 8  \\\cline{3-4}
		& \multicolumn{1}{c}{} & \multicolumn{2}{c}{State $2$B} \\
	\end{tabular}
    \caption{Payoff matrices of the two-step game after the Agent 1 chose the first action. Action A takes the agents to State $2$A and action B takes them to State $2$B.}
    \label{tab:2step_game}
\end{table}

\begin{table}[h]
	\setlength{\extrarowheight}{3pt}
	\centering
	(a)
	\begin{tabular}{c|*{2}{>{\centering\arraybackslash}p{.08\linewidth}|}}
		\multicolumn{1}{c}{} & \multicolumn{2}{c}{State $1$} \\
		\multicolumn{1}{c}{} & \multicolumn{1}{c}{\bb{$A$}}  & \multicolumn{1}{c}{\bb{$B$}} \\ \cline{2-3}
		\cc{$A$} & 6.94 & 6.94 \\ \cline{2-3}
		\cc{$B$} & 6.35 & 6.36  \\\cline{2-3}
	\end{tabular}~
	\begin{tabular}{|*{2}{>{\centering\arraybackslash}p{.08\linewidth}|}}
		\multicolumn{2}{c}{State $2$A} \\
		\multicolumn{1}{c}{\bb{$A$}}  & \multicolumn{1}{c}{\bb{$B$}} \\ \cline{1-2}
		6.99 & 7.02 \\\cline{1-2}
		6.99 & 7.02  \\\cline{1-2}
	\end{tabular}~
	\begin{tabular}{|*{2}{>{\centering\arraybackslash}p{.08\linewidth}|}}
		\multicolumn{2}{c}{State $2$B} \\
		\multicolumn{1}{c}{\bb{$A$}}  & \multicolumn{1}{c}{\bb{$B$}} \\\cline{1-2}
		\text{-1.87} & 2.31 \\\cline{1-2}
		2.33 & 6.51  \\\cline{1-2}
	\end{tabular}\\\bigskip

	(b)
	\begin{tabular}{c|*{2}{>{\centering\arraybackslash}p{.08\linewidth}|}}
		\multicolumn{1}{c}{} & \multicolumn{1}{c}{\bb{$A$}}  & \multicolumn{1}{c}{\bb{$B$}} \\ \cline{2-3}
		\cc{$A$} & 6.93 & 6.93  \\ \cline{2-3}
		\cc{$B$} & 7.92 & 7.92  \\\cline{2-3}
	\end{tabular}~
	\begin{tabular}{|*{2}{>{\centering\arraybackslash}p{.08\linewidth}|}}
		\multicolumn{1}{c}{\bb{$A$}}  & \multicolumn{1}{c}{\bb{$B$}} \\ \cline{1-2}
		7.00 & 7.00 \\ \cline{1-2}
		7.00 & 7.00  \\\cline{1-2}
	\end{tabular}~
	\begin{tabular}{|*{2}{>{\centering\arraybackslash}p{.08\linewidth}|}}
		\multicolumn{1}{c}{\bb{$A$}}  & \multicolumn{1}{c}{\bb{$B$}} \\\cline{1-2}
		0.00 & 1.00 \\\cline{1-2}
		1.00 & 8.00 \\\cline{1-2}
	\end{tabular}\\
    \caption{$Q_{tot}$ on the two-step game for (a) VDN and (b) QMIX.}
    \label{qmix_2step_game_main}
\end{table}

Table \ref{qmix_2step_game_main}, which shows the learned values for $Q_{tot}$, demonstrates that QMIX's higher representational capacity allows it to accurately represent the joint-action value function whereas VDN cannot. This directly translates into VDN learning the suboptimal strategy of selecting Action A at the first step and receiving a reward of 7, whereas QMIX recovers the optimal strategy from its learnt joint-action values and receives a reward of 8. 

\begin{figure*}[htb!]
    \centering
    \includegraphics[height=0.3cm]{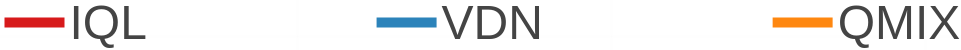}
    \vfill
    \subfigure[3m]{
        \includegraphics[width=0.315\textwidth]{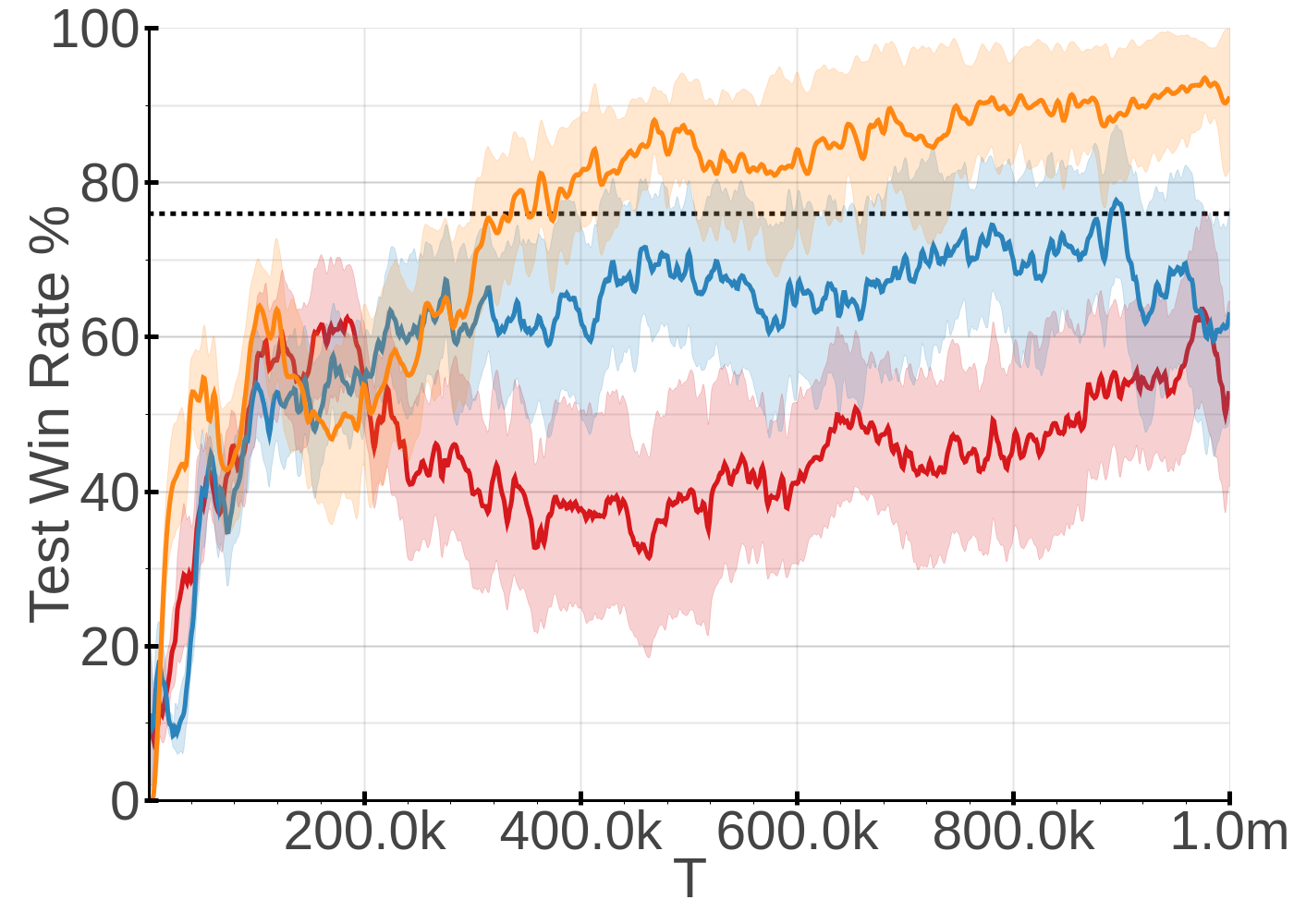}
    }
    \subfigure[5m]{
        \includegraphics[width=0.315\textwidth]{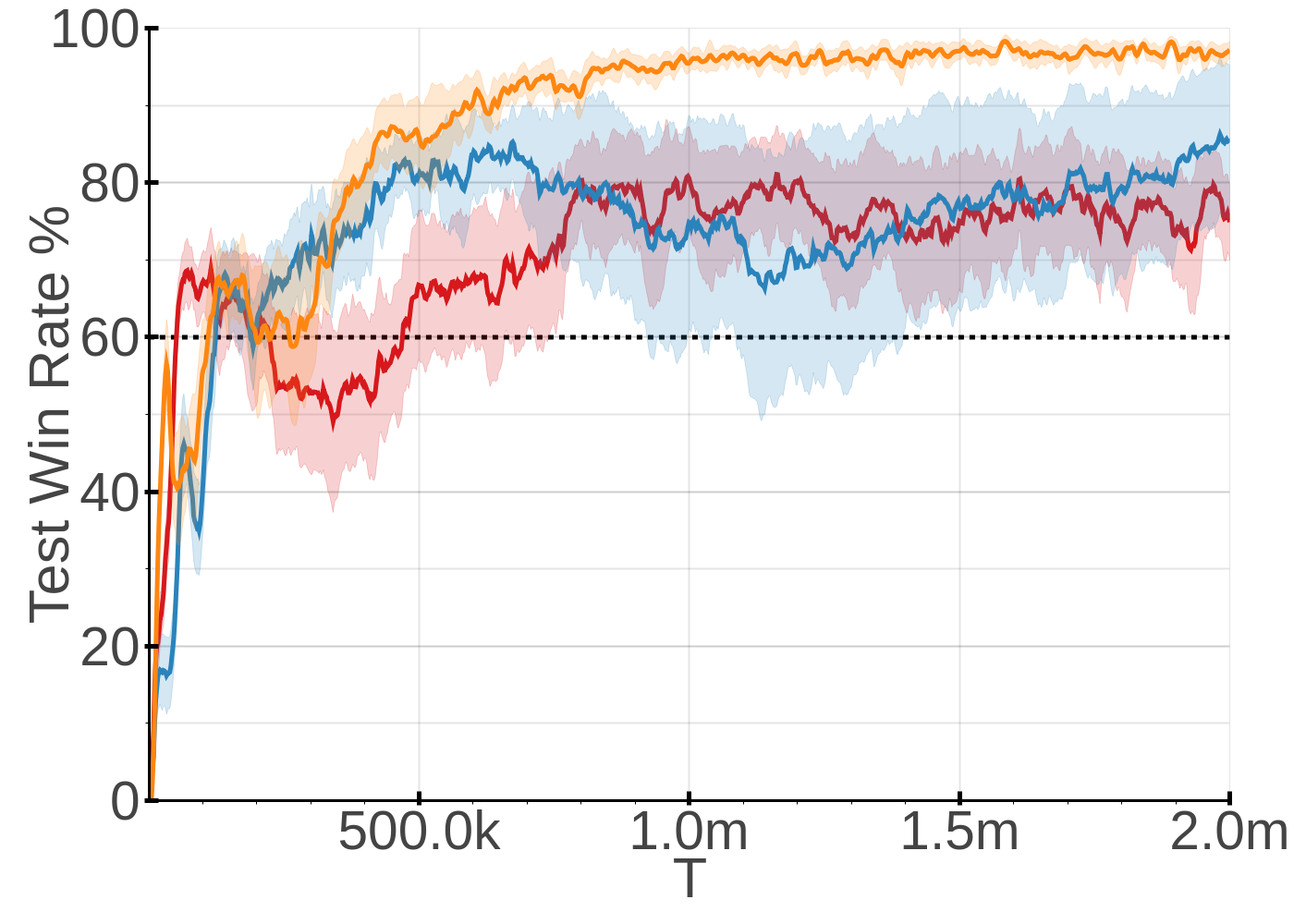}
    }
    \subfigure[8m]{
        \includegraphics[width=0.315\textwidth]{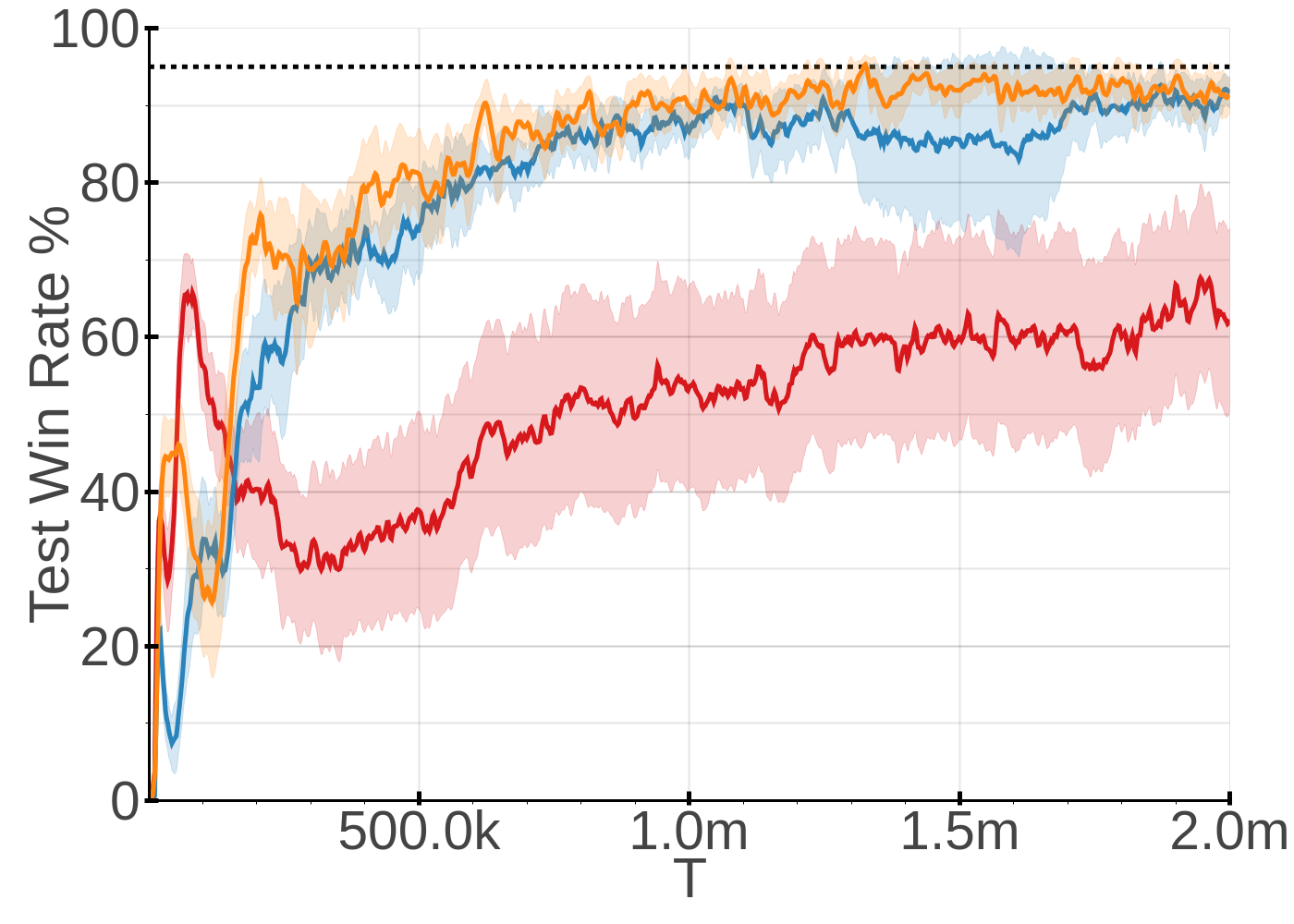}
    }
    \\
    \subfigure[2s\_3z]{
        \includegraphics[width=0.315\textwidth]{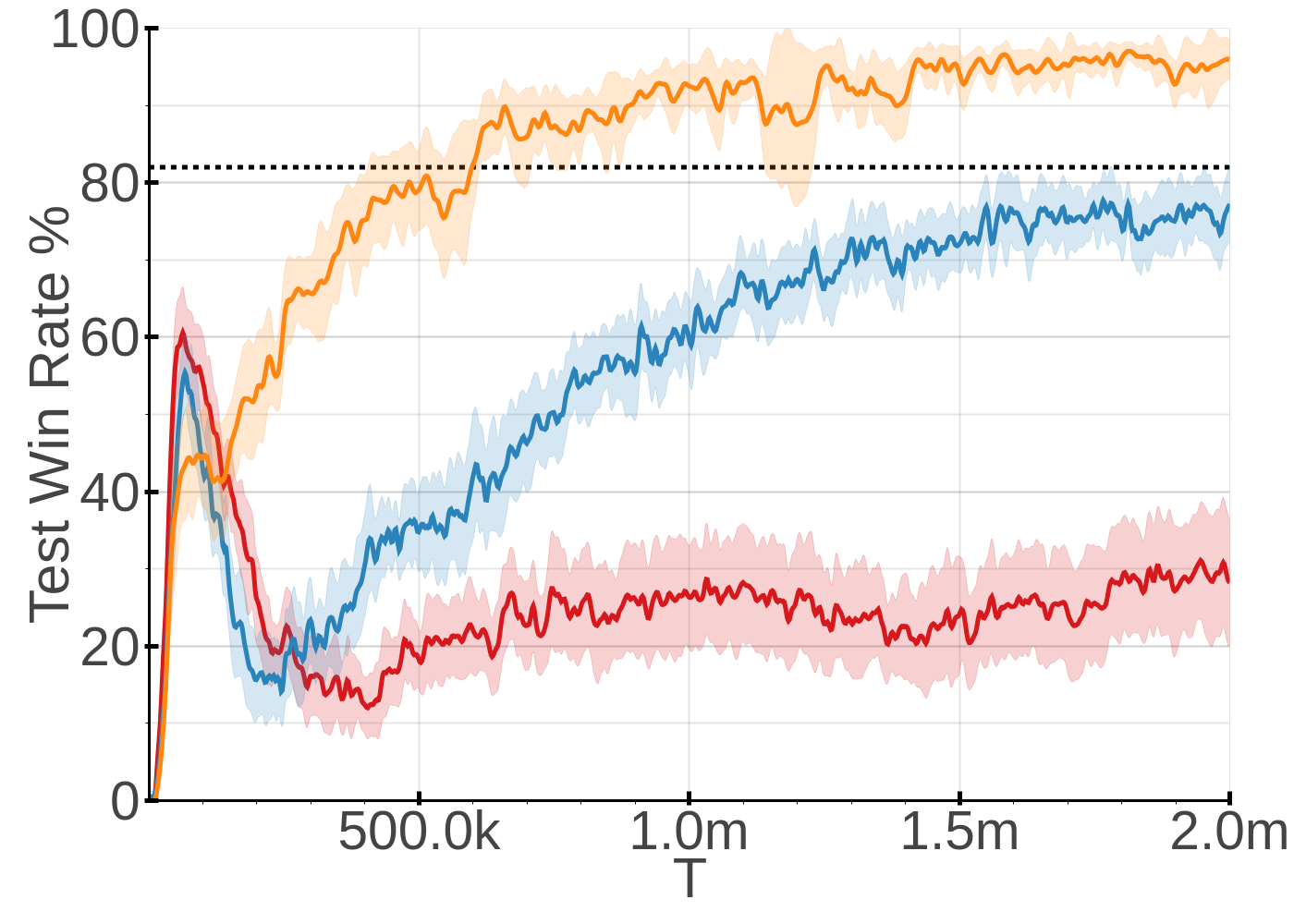}
    }
    \subfigure[3s\_5z]{
        \includegraphics[width=0.315\textwidth]{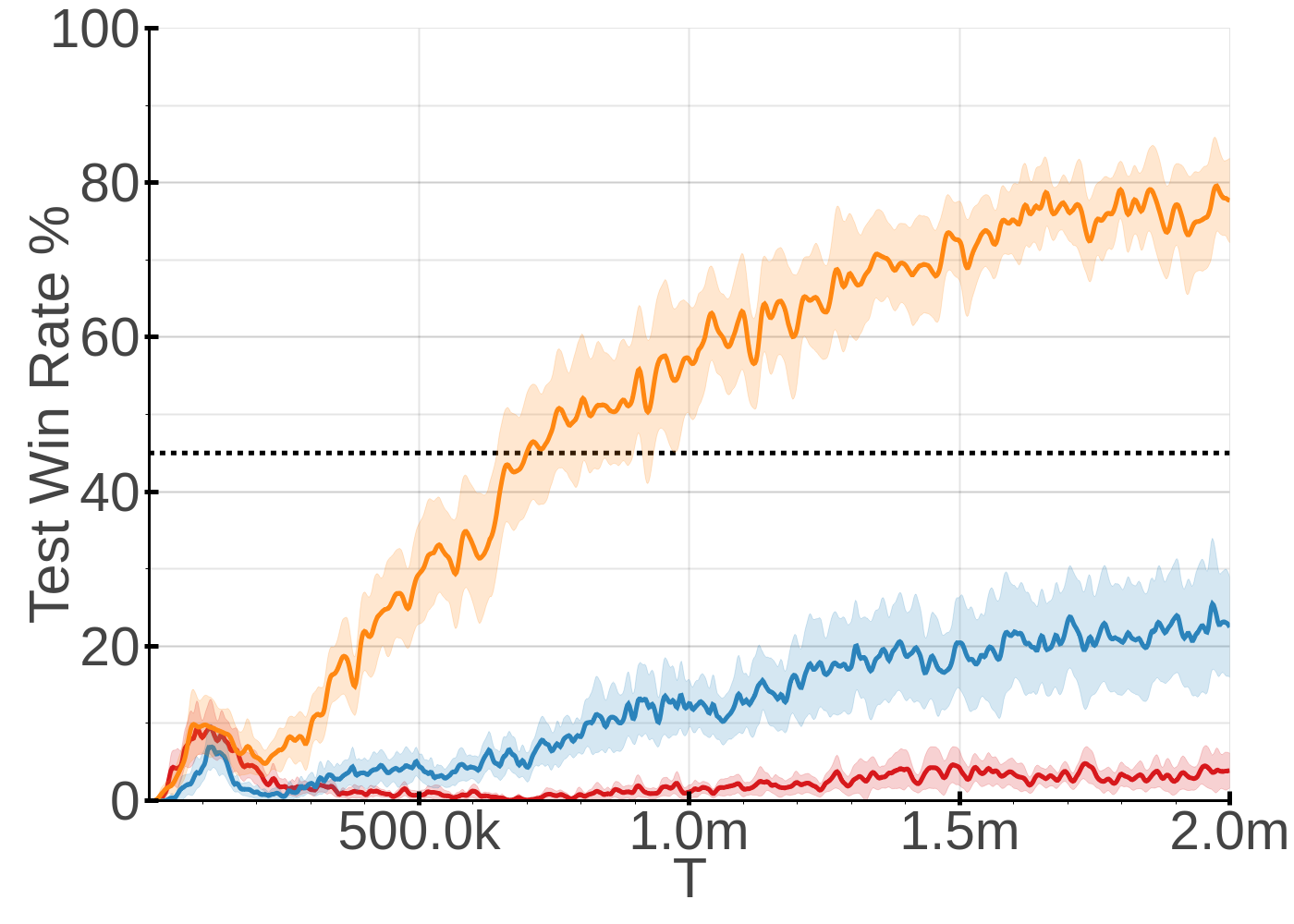}
    }
    \subfigure[1c\_3s\_5z]{
        \includegraphics[width=0.315\textwidth]{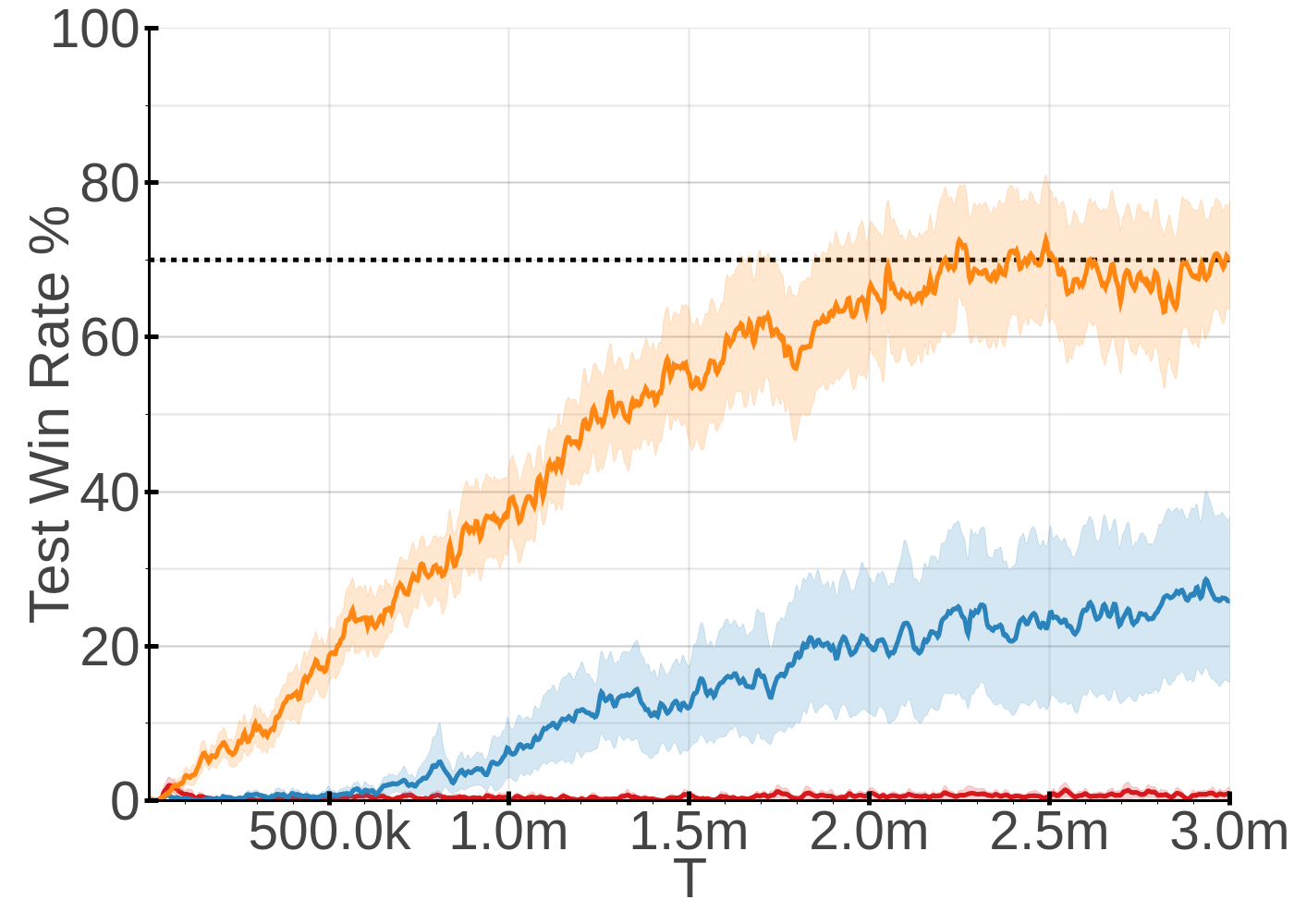}
    }
    \caption{Win rates for IQL, VDN, and QMIX on six different combat maps. The performance of the heuristic-based algorithm is shown as a dashed line.}
    \label{fig:starcraft_res}
\end{figure*}

\section{Experimental Setup}
\label{sec:setting}

In this section, we describe the decentralised StarCraft II micromanagement problems to which we apply QMIX and the ablations we consider.

\subsection{Decentralised StarCraft II Micromanagement} 
Real-time strategy (RTS) games have recently emerged as challenging benchmarks for the RL community. 
StarCraft, in particular, offers a great opportunity to tackle competitive and cooperative multi-agent problems.
Units in StarCraft have a rich set of complex micro-actions that allow the learning of complex interactions between collaborating agents. 
Previous work \cite{usunier_episodic_2016, foerster_counterfactual_2017, peng_multiagent_2017} applied RL to the original version of StarCraft: BW, which made use of the standard API or related wrappers \cite{synnaeve_torchcraft_2016}. 
We perform our experiments on the StarCraft II Learning Environment (SC2LE) \cite{vinyals_starcraft_2017}, which is based on the second version of the game. 
Because it is supported by the developers of the game, SC2LE mitigates many of the practical difficulties in using StarCraft as an RL platform, such as the dependence on complicated APIs and external emulation software.

In this work, we focus on the \textit{decentralised micromanagement} problem in StarCraft II, in which each of the learning agents controls an individual army unit. 
We consider combat scenarios where two groups of identical units are placed symmetrically on the map. 
The units of the first, allied, group are controlled by the decentralised agents.  
The enemy units are controlled by a built-in StarCraft II AI, which makes use of handcrafted heuristics. 
The initial placement of units within the groups varies across episodes.  
The difficulty of the computer AI controlling the enemy units is set to \texttt{medium}.
At the beginning of each episode, the enemy units are ordered to attack the allies.
We compare our results on a set of maps where each unit group consists of 3 Marines (3m), 5 Marines (5m), 8 Marines (8m), 2 Stalkers and 3 Zealots (2s\_3z), 3 Stalkers and 5 Zealots (3s\_5z), or 1 Colossus, 3 Stalkers and 5 Zealots (1c\_3s\_5z).

Similar to the work of \citet{foerster_counterfactual_2017}, the action space of agents consists of the following set of discrete actions: \texttt{move[direction]}, \texttt{attack[enemy\_id]}, \texttt{stop}, and
\texttt{noop}. 
Agents can only move in four directions: north, south, east, or west.
A unit is allowed to perform the \texttt{attack[enemy\_id]} action only if the 
enemy is within its \textit{shooting range}. 
This facilitates the decentralisation of the problem and prohibits the usage of the \emph{attack-move} macro-actions that are integrated into the game. 
Furthermore, we disable the following unit behaviour when idle: responding to enemy fire and attacking enemies if they are in range. 
By doing so, we force the agents to explore in order to find the optimal combat strategy themselves, rather than relying on built-in StarCraft II utilities.

Partial observability is achieved by the introduction of unit \textit{sight range}, which restricts the agents from receiving information about allied or enemy units that are out of range. 
Moreover, agents can only observe others if they are alive and cannot distinguish between units that are dead or out of range.

At each time step, the agents receive a joint reward equal to the total damage dealt on the enemy units. 
In addition, agents receive a bonus of $10$ points after killing each opponent, and $200$ points after killing all opponents. These rewards are all normalised to ensure the maximum cumulative reward achievable in an episode is $20$.

The full details of the environmental setup, architecture and training are available in the supplementary material.

\subsection{Ablations} 
\label{sub:abl}

We perform ablation experiments in order to investigate the influence of the inclusion of extra state information and the necessity of non-linear transformations in the mixing network. 

First, we analyse the significance of extra state information on the mixing network by comparing against QMIX without hypernetworks. Thus, the weights and biases of the mixing network are learned in the standard way, without conditioning on the state. We refer to this method as QMIX-NS. We take the absolute value of the weights in order to enforce the monotonicity constraint.

Second, we investigate the necessity of non-linear mixing by removing the hidden layer of the mixing network. This method can be thought of as an extension of VDN that uses the state $s$ to perform a weighted sum over $Q_a$ values. We call this method QMIX-Lin.

Third, we investigate the significance of utilising the state $s$ in comparison to the non-linear mixing. To do this we extend VDN by adding a state-dependent term to the sum of the agent's $Q$-Values. This state-dependent term is produced by a network with a single hidden layer of $32$ units and a ReLU non-linearity, taking in the state $s$ as input (the same as the hypernetwork producing the final bias in QMIX). We refer to this method as VDN-S.

We also show the performance of a non-learning heuristic-based algorithm with full observability, where each agent attacks the closest enemy and continues attacking the same target until the unit dies. Afterwards, the agent starts attacking the nearest enemy and so forth.

 \section{Results}
\label{sec:results}

\begin{figure*}[htb!]
    \centering
    \includegraphics[height=0.3cm]{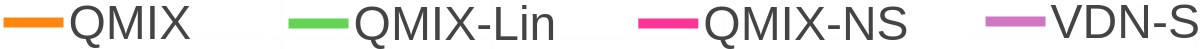}
    \vfill
    \subfigure[3m]{
        \includegraphics[width=0.31\textwidth]{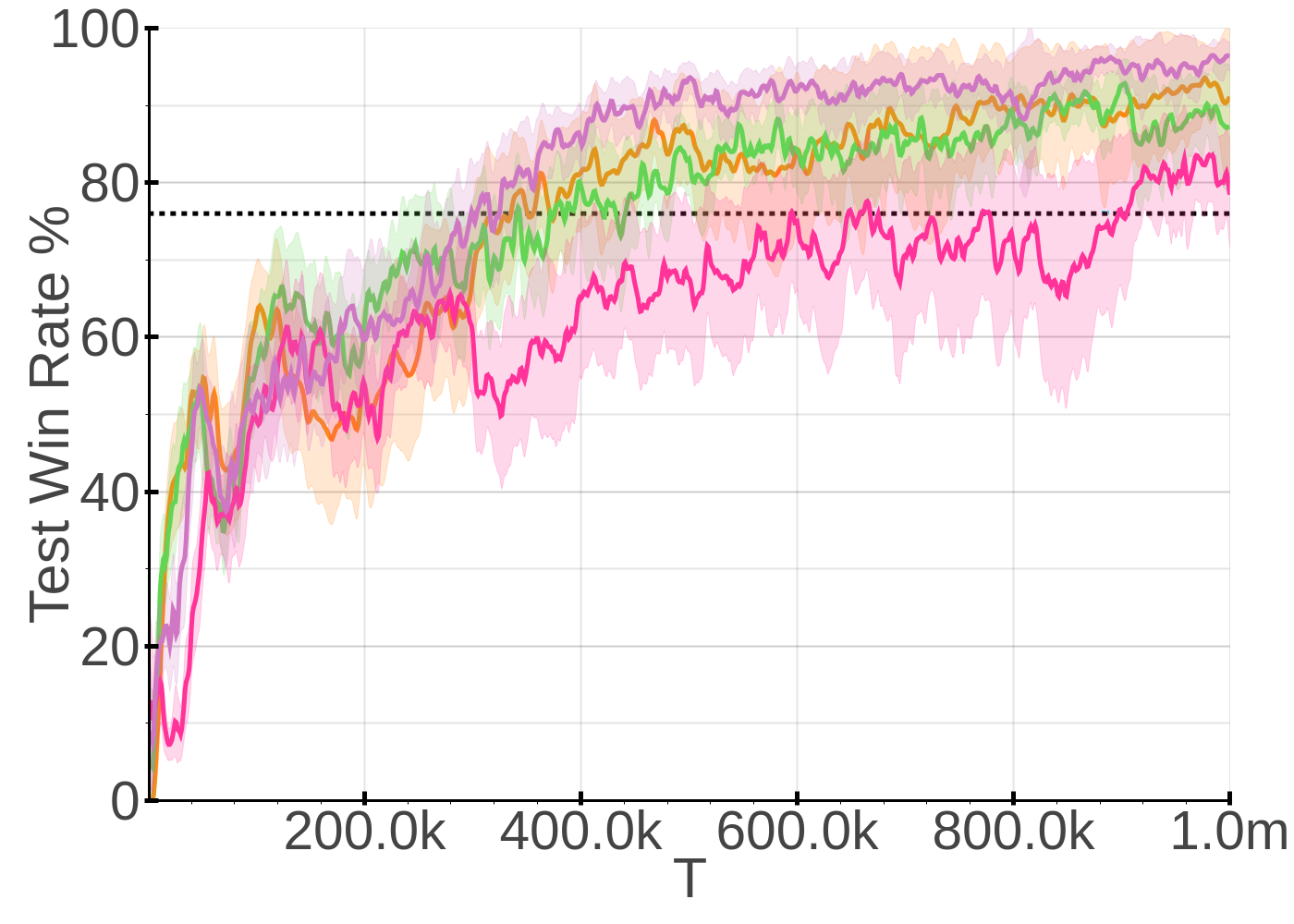}
    }
    \subfigure[2s\_3z]{
        \includegraphics[width=0.31\textwidth]{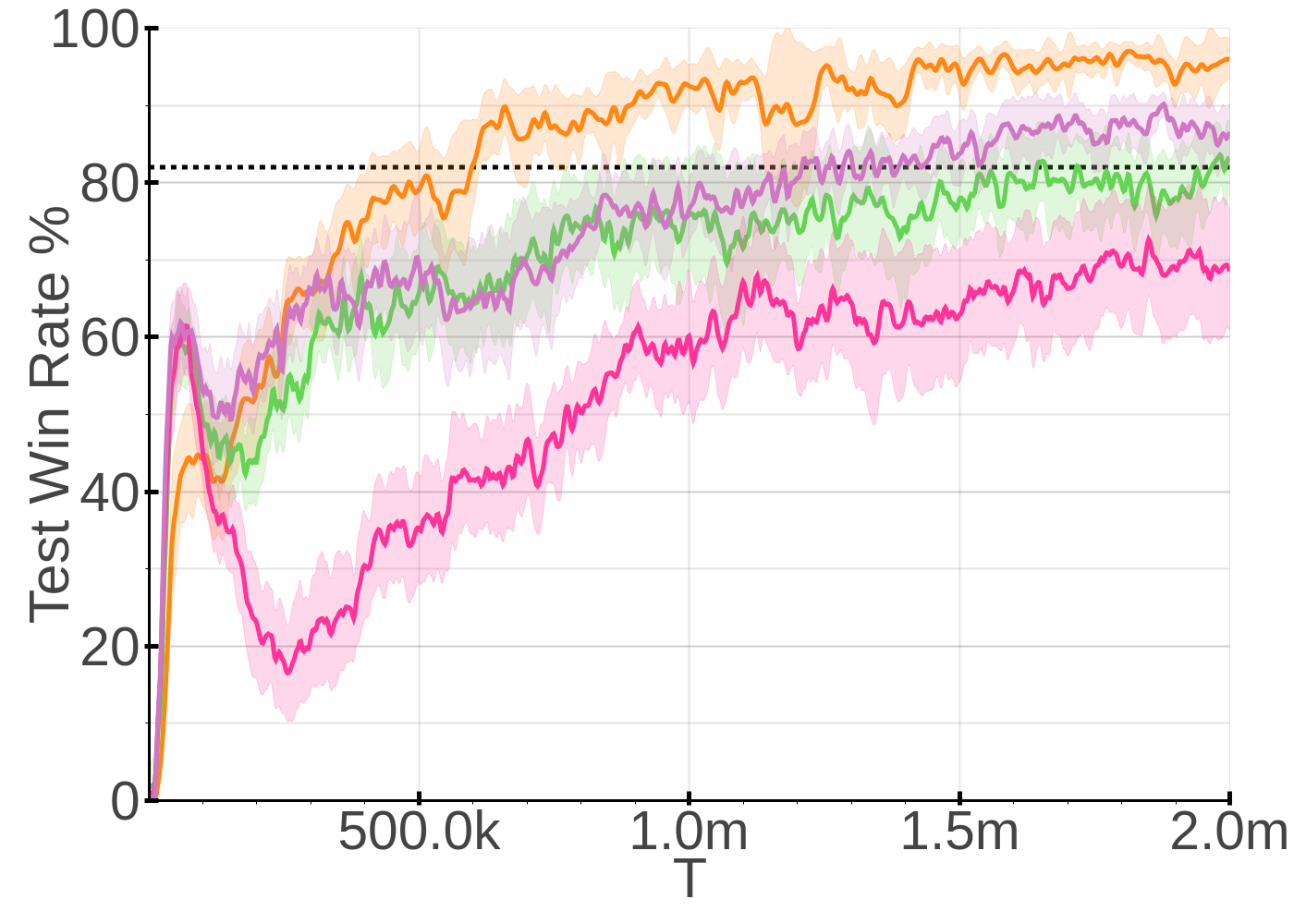}
    }
    \subfigure[3s\_5z]{
        \includegraphics[width=0.31\textwidth]{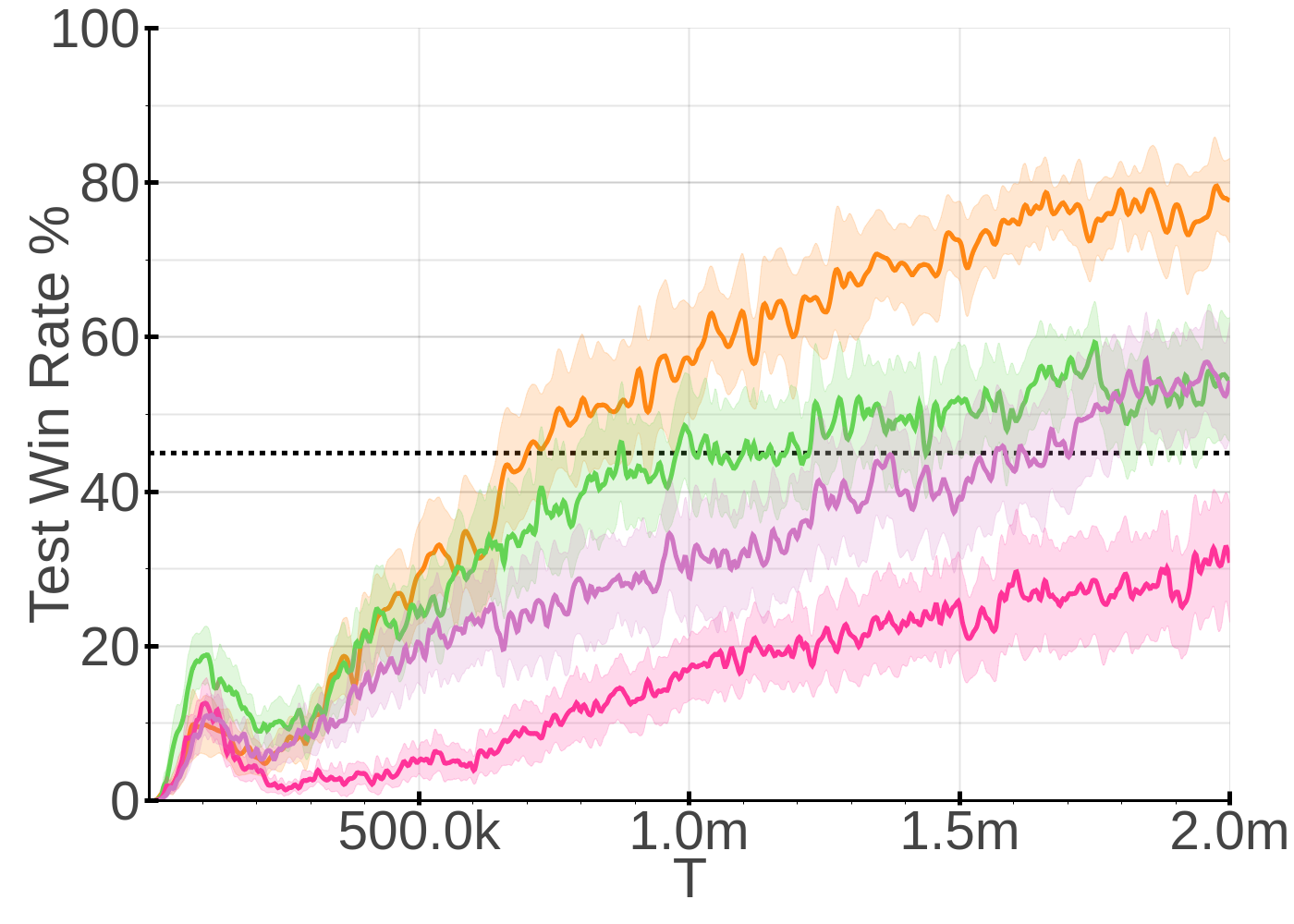}
    }
    \caption{Win rates for QMIX and ablations on 3m, 2s\_3z and 3s\_5z maps.}
    \label{fig:starcraft_ablations}
\end{figure*}

In order to evaluate each method's performance, we adopt the following evaluation procedure: for each run of a method, we pause training every 100 episodes and run 20 independent episodes with each agent performing greedy decentralised action selection. The percentage of these episodes in which the method defeats all enemy units within the time limit is referred to as the \emph{test win rate}. 

Figures \ref{fig:starcraft_res} and \ref{fig:starcraft_ablations} plot the mean test win rate across 20 runs for each method on selected maps, together with 95$\%$ confidence intervals. The graphs for all methods on all maps are available in the supplementary material.

\subsection{Main Results}
In all scenarios, IQL fails to learn a policy that consistently defeats the enemy. In addition, the training is highly unstable due to the non-stationarity of the environment which arises due to the other agents changing their behaviour during training. 

The benefits of learning the joint action-value function can be demonstrated by VDN's superior performance over IQL in all scenarios. VDN is able to more consistently learn basic coordinated behaviour, in the form of \textit{focus firing} which allows it to win the majority of its encounters on the 5m and 8m maps. On the 8m map, this simple strategy is sufficient for good performance, as evidenced by the extremely high win rate of the heuristic-based algorithm, and explains the performance parity with QMIX. However, on the 3m task, which requires more fine-grained control, it is unable to learn to consistently defeat the enemy.

QMIX is noticeably the strongest performer on all of the maps, in particular on the maps with hetergenous agent types. The largest performance gap can be seen in the 3s\_5z and 1c\_3s\_5z maps, where VDN is unable to reach the performance of the simple heuristic. The superior representational capacity of QMIX combined with the state information presents a clear benefit over a more restricted linear decomposition. 

\subsection{Ablation Results}

Our additional ablation experiments reveal that QMIX outperforms, or is competitive with, all of its ablations discussed in Section \ref{sub:abl}. Figure \ref{fig:starcraft_ablations}a shows that non-linear value function factorisation is not always required on a map with homogeneous agent types. However, the additional complexity introduced through the extra hidden layer does not slow down learning. In contrast, Figures \ref{fig:starcraft_ablations}b and \ref{fig:starcraft_ablations}c show that on a map with heterogeneous agent types a combination of both central state information and non-linear value function factorisation is required to achieve good performance. QMIX-NS performs on par or slightly better than VDN in both scenarios, which suggests that a non-linear decomposition is not always beneficial when not conditioning on the central state in complex scenarios. Additionally, the performance of VDN-S compared to QMIX-Lin shows the necessity of allowing a non-linear mixing in order to fully leverage central state information.

\subsection{Learned Policies}

We examine the learned behaviours of the policies in order to better understand the differences between the strategies learnt by the different methods. 
On the 8m scenario, both QMIX and VDN learn the particularly sophisticated strategy of first positioning the units into a semicircle in order to fire at the incoming enemy units from the sides (as opposed to just head on).
On the 2s\_3z scenario, VDN first runs left and then attacks the enemy once they are in range with no regards to positioning or unit match-ups. QMIX, on the other hand learns to position the Stalkers so that the enemy Zealots cannot attack them. This is especially important since Zealots \emph{counter} Stalkers. QMIX achieves this by having the allied Zealots first block off and then attack the enemy Zealots (whilst the Stalkers fire from a safe distance), before moving on to the enemy Stalkers. The same behaviour is observed in the 3s\_5z scenario for QMIX. VDN-S does not learn to protect the Stalkers from the Zealots, and first positions the units around their starting location and then attacks the enemy as they move in.

The initial hump in the performance of both VDN and IQL is due to both methods initially learning the simple strategy of just attacking the first visible enemy (which is quite successful as shown by the heuristic). However, due to exploratory learning behaviour, they also attempt to move around (instead of just firing), which results in the rapid decline in performance. IQL is unable to then recover the initial strategy, whereas VDN  learns how to combine small movements and firing together. \section{Conclusion}

This paper presented QMIX, a deep multi-agent RL method that allows end-to-end learning of decentralised policies in a centralised setting and makes efficient use of extra state information. QMIX allows the learning of a rich joint action-value function, which admits tractable decompositions into per-agent action-value functions. This is achieved by imposing a monotonicity constraint on the mixing network.

Our results in decentralised unit micromanagement tasks in StarCraft II show that QMIX improves the final performance over other value-based multi-agent methods that employ less sophisticated joint state-value function factorisation, as well as independent $Q$-learning. 

In the near future, we aim to conduct additional experiments to compare the methods across tasks with a larger number and greater diversity of units. In the longer term, we aim to complement QMIX with more coordinated exploration schemes for settings with many learning agents. 
 \section*{Acknowledgements} 

This project has received funding from the European Research Council (ERC) under the European Union's Horizon 2020 research and innovation programme (grant agreement number 637713). 
It was also supported by the Oxford-Google DeepMind Graduate Scholarship, the UK EPSRC CDT in Autonomous Intelligent Machines and Systems, Chevening Scholarship, Luys Scholarship and an EPSRC grant (EP/M508111/1, EP/N509711/1). This work is linked to and partly funded by the project Free the Drones (FreeD) under the Innovation Fund Denmark and Microsoft. The experiments were made possible by a generous equipment grant from NVIDIA.

We would like to thank Frans Oliehoek and Wendelin Boehmer for helpful comments and discussion. 
We also thank Oriol Vinyals, Kevin Calderone, and the rest of the SC2LE team at DeepMind and Blizzard for their work on the interface.

\bibliography{final}

\begin{thebibliography}{32}
\providecommand{\natexlab}[1]{#1}
\providecommand{\url}[1]{\texttt{#1}}
\expandafter\ifx\csname urlstyle\endcsname\relax
  \providecommand{\doi}[1]{doi: #1}\else
  \providecommand{\doi}{doi: \begingroup \urlstyle{rm}\Url}\fi

\bibitem[Busoniu et~al.(2008)Busoniu, Babuska, and
  De~Schutter]{busoniu_comprehensive_2008}
Busoniu, L., Babuska, R., and De~Schutter, B.
\newblock A {Comprehensive} {Survey} of {Multiagent} {Reinforcement}
  {Learning}.
\newblock \emph{IEEE Transactions on Systems, Man, and Cybernetics, Part C
  (Applications and Reviews)}, 38\penalty0 (2):\penalty0 156--172, 2008.

\bibitem[Cao et~al.(2012)Cao, Yu, Ren, and Chen]{cao_overview_2012}
Cao, Y., Yu, W., Ren, W., and Chen, G.
\newblock An {Overview} of {Recent} {Progress} in the {Study} of {Distributed}
  {Multi}-agent {Coordination}.
\newblock \emph{IEEE Transactions on Industrial Informatics}, 9\penalty0
  (1):\penalty0 427--438, 2012.

\bibitem[Chung et~al.(2014)Chung, Gulcehre, Cho, and
  Bengio]{chung_empirical_2014}
Chung, J., Gulcehre, C., Cho, K., and Bengio, Y.
\newblock Empirical evaluation of gated recurrent neural networks on sequence
  modeling.
\newblock In \emph{NIPS 2014 Workshop on Deep Learning}, 2014.

\bibitem[Dugas et~al.(2009)Dugas, Bengio, Blisle, Nadeau, and
  Garcia]{Dugas_2009}
Dugas, C., Bengio, Y., Blisle, F., Nadeau, C., and Garcia, R.
\newblock Incorporating functional knowledge in neural networks.
\newblock \emph{Journal of Machine Learning Research}, 10:\penalty0 1239--1262,
  2009.

\bibitem[Foerster et~al.(2017)Foerster, Nardelli, Farquhar, Afouras, Torr,
  Kohli, and Whiteson]{foerster_stabilising_2017}
Foerster, J., Nardelli, N., Farquhar, G., Afouras, T., Torr, P. H.~S., Kohli,
  P., and Whiteson, S.
\newblock Stabilising {Experience} {Replay} for {Deep} {Multi}-{Agent}
  {Reinforcement} {Learning}.
\newblock In \emph{Proceedings of {The} 34th {International} {Conference} on
  {Machine} {Learning}}, pp.\  1146--1155, 2017.

\bibitem[Foerster et~al.(2018)Foerster, Farquhar, Afouras, Nardelli, and
  Whiteson]{foerster_counterfactual_2017}
Foerster, J., Farquhar, G., Afouras, T., Nardelli, N., and Whiteson, S.
\newblock Counterfactual multi-agent policy gradients.
\newblock In \emph{Proceedings of the Thirty-Second AAAI Conference on
  Artificial Intelligence}, 2018.

\bibitem[Guestrin et~al.(2002)Guestrin, Koller, and
  Parr]{guestrin_multiagent_2002}
Guestrin, C., Koller, D., and Parr, R.
\newblock Multiagent {Planning} with {Factored} {MDPs}.
\newblock In \emph{Advances in {Neural} {Information} {Processing} {Systems}},
  pp.\  1523--1530. MIT Press, 2002.

\bibitem[Gupta et~al.(2017)Gupta, Egorov, and
  Kochenderfer]{gupta_cooperative_2017}
Gupta, J.~K., Egorov, M., and Kochenderfer, M.
\newblock Cooperative {Multi}-agent {Control} {Using} {Deep} {Reinforcement}
  {Learning}.
\newblock In \emph{Autonomous {Agents} and {Multiagent} {Systems}}, pp.\
  66--83. Springer, 2017.

\bibitem[Ha et~al.(2017)Ha, Dai, and Le]{ha_hypernetworks_2016}
Ha, D., Dai, A., and Le, Q.~V.
\newblock {HyperNetworks}.
\newblock In \emph{Proceedings of the International Conference on Learning
  Representations (ICLR)}, 2017.

\bibitem[Hausknecht \& Stone(2015)Hausknecht and Stone]{hausknecht_deep_2015}
Hausknecht, M. and Stone, P.
\newblock Deep {Recurrent} {Q}-{Learning} for {Partially} {Observable} {MDPs}.
\newblock In \emph{AAAI Fall Symposium on Sequential Decision Making for
  Intelligent Agents}, 2015.

\bibitem[Hochreiter \& Schmidhuber(1997)Hochreiter and
  Schmidhuber]{hochreiter_long_1997}
Hochreiter, S. and Schmidhuber, J.
\newblock Long short-term memory.
\newblock \emph{Neural computation}, 9\penalty0 (8):\penalty0 1735--1780, 1997.

\bibitem[H{\"u}ttenrauch et~al.(2017)H{\"u}ttenrauch, {\v S}o{\v s}i{\'c}, and
  Neumann]{huttenrauch_guided_2017}
H{\"u}ttenrauch, M., {\v S}o{\v s}i{\'c}, A., and Neumann, G.
\newblock Guided {Deep} {Reinforcement} {Learning} for {Swarm} {Systems}.
\newblock In \emph{AAMAS 2017 Autonomous Robots and Multirobot Systems (ARMS)
  Workshop}, 2017.

\bibitem[Jorge et~al.(2016)Jorge, K{\aa}geb{\"a}ck, and
  Gustavsson]{jorge_learning_2016}
Jorge, E., K{\aa}geb{\"a}ck, M., and Gustavsson, E.
\newblock Learning to play guess who? and inventing a grounded language as a
  consequence.
\newblock In \emph{NIPS 2016 Workshop on Deep Reinforcement Learning}, 2016.

\bibitem[Kok \& Vlassis(2006)Kok and Vlassis]{kok_collaborative_2006}
Kok, J.~R. and Vlassis, N.
\newblock Collaborative {Multiagent} {Reinforcement} {Learning} by {Payoff}
  {Propagation}.
\newblock \emph{Journal of Machine Learning Research}, 7:\penalty0 1789--1828,
  2006.

\bibitem[Kraemer \& Banerjee(2016)Kraemer and
  Banerjee]{kraemer_multi-agent_2016}
Kraemer, L. and Banerjee, B.
\newblock Multi-agent reinforcement learning as a rehearsal for decentralized
  planning.
\newblock \emph{Neurocomputing}, 190:\penalty0 82--94, 2016.

\bibitem[Leibo et~al.(2017)Leibo, Zambaldi, Lanctot, Marecki, and
  Graepel]{leibo_multi-agent_2017}
Leibo, J.~Z., Zambaldi, V., Lanctot, M., Marecki, J., and Graepel, T.
\newblock Multi-agent reinforcement learning in sequential social dilemmas.
\newblock In \emph{Proceedings of the 16th Conference on Autonomous Agents and
  Multiagent Systems}, pp.\  464--473, 2017.

\bibitem[Lowe et~al.(2017)Lowe, Wu, Tamar, Harb, Abbeel, and
  Mordatch]{lowe_multi-agent_2017}
Lowe, R., Wu, Y., Tamar, A., Harb, J., Abbeel, O.~P., and Mordatch, I.
\newblock Multi-agent actor-critic for mixed cooperative-competitive
  environments.
\newblock In \emph{Advances in Neural Information Processing Systems}, pp.\
  6382--6393, 2017.

\bibitem[Mnih et~al.(2015)Mnih, Kavukcuoglu, Silver, Rusu, Veness, Bellemare,
  Graves, Riedmiller, Fidjeland, Ostrovski, and
  {others}]{mnih_human-level_2015}
Mnih, V., Kavukcuoglu, K., Silver, D., Rusu, A.~A., Veness, J., Bellemare,
  M.~G., Graves, A., Riedmiller, M., Fidjeland, A.~K., Ostrovski, G., and
  {others}.
\newblock Human-level control through deep reinforcement learning.
\newblock \emph{Nature}, 518\penalty0 (7540):\penalty0 529--533, 2015.

\bibitem[Oliehoek \& Amato(2016)Oliehoek and Amato]{oliehoek_concise_2016}
Oliehoek, F.~A. and Amato, C.
\newblock \emph{A {Concise} {Introduction} to {Decentralized} {POMDPs}}.
\newblock {SpringerBriefs} in {Intelligent} {Systems}. Springer, 2016.

\bibitem[Oliehoek et~al.(2008)Oliehoek, Spaan, and
  Vlassis]{oliehoek_optimal_2008}
Oliehoek, F.~A., Spaan, M. T.~J., and Vlassis, N.
\newblock Optimal and {Approximate} {Q}-value {Functions} for {Decentralized}
  {POMDPs}.
\newblock \emph{Journal of Artificial Intelligence Research}, 32\penalty0
  (1):\penalty0 289--353, 2008.

\bibitem[Omidshafiei et~al.(2017)Omidshafiei, Pazis, Amato, How, and
  Vian]{omidshafiei_deep_2017}
Omidshafiei, S., Pazis, J., Amato, C., How, J.~P., and Vian, J.
\newblock Deep {Decentralized} {Multi}-task {Multi}-{Agent} {RL} under
  {Partial} {Observability}.
\newblock In \emph{Proceedings of the 34th International Conference on Machine
  Learning}, pp.\  2681--2690, 2017.

\bibitem[Peng et~al.(2017)Peng, Wen, Yang, Yuan, Tang, Long, and
  Wang]{peng_multiagent_2017}
Peng, P., Wen, Y., Yang, Y., Yuan, Q., Tang, Z., Long, H., and Wang, J.
\newblock Multiagent {Bidirectionally}-{Coordinated} {Nets}: {Emergence} of
  {Human}-level {Coordination} in {Learning} to {Play} {StarCraft} {Combat}
  {Games}.
\newblock \emph{arXiv preprint arXiv:1703.10069}, 2017.

\bibitem[Pinkus(1999)]{pinkus1999approximation}
Pinkus, A.
\newblock Approximation theory of the mlp model in neural networks.
\newblock \emph{Acta numerica}, 8:\penalty0 143--195, 1999.

\bibitem[Sukhbaatar et~al.(2016)Sukhbaatar, Fergus, and
  {others}]{sukhbaatar_learning_2016}
Sukhbaatar, S., Fergus, R., and {others}.
\newblock Learning multiagent communication with backpropagation.
\newblock In \emph{Advances in {Neural} {Information} {Processing} {Systems}},
  pp.\  2244--2252, 2016.

\bibitem[Sunehag et~al.(2017)Sunehag, Lever, Gruslys, Czarnecki, Zambaldi,
  Jaderberg, Lanctot, Sonnerat, Leibo, Tuyls, and
  Graepel]{sunehag_value-decomposition_2017}
Sunehag, P., Lever, G., Gruslys, A., Czarnecki, W.~M., Zambaldi, V., Jaderberg,
  M., Lanctot, M., Sonnerat, N., Leibo, J.~Z., Tuyls, K., and Graepel, T.
\newblock Value-{Decomposition} {Networks} {For} {Cooperative} {Multi}-{Agent}
  {Learning} {Based} {On} {Team} {Reward}.
\newblock In \emph{Proceedings of the 17th International Conference on
  Autonomous Agents and Multiagent Systems}, 2017.

\bibitem[Synnaeve et~al.(2016)Synnaeve, Nardelli, Auvolat, Chintala, Lacroix,
  Lin, Richoux, and Usunier]{synnaeve_torchcraft_2016}
Synnaeve, G., Nardelli, N., Auvolat, A., Chintala, S., Lacroix, T., Lin, Z.,
  Richoux, F., and Usunier, N.
\newblock {TorchCraft}: a {Library} for {Machine} {Learning} {Research} on
  {Real}-{Time} {Strategy} {Games}.
\newblock \emph{arXiv preprint arXiv:1611.00625}, 2016.

\bibitem[Tampuu et~al.(2017)Tampuu, Matiisen, Kodelja, Kuzovkin, Korjus, Aru,
  Aru, and Vicente]{tampuu_multiagent_2015}
Tampuu, A., Matiisen, T., Kodelja, D., Kuzovkin, I., Korjus, K., Aru, J., Aru,
  J., and Vicente, R.
\newblock Multiagent cooperation and competition with deep reinforcement
  learning.
\newblock \emph{PloS one}, 2017.

\bibitem[Tan(1993)]{tan_multi-agent_1993}
Tan, M.
\newblock Multi-agent reinforcement learning: {Independent} vs. cooperative
  agents.
\newblock In \emph{Proceedings of the Tenth International Conference on Machine
  Learning}, pp.\  330--337, 1993.

\bibitem[Usunier et~al.(2017)Usunier, Synnaeve, Lin, and
  Chintala]{usunier_episodic_2016}
Usunier, N., Synnaeve, G., Lin, Z., and Chintala, S.
\newblock Episodic {Exploration} for {Deep} {Deterministic} {Policies}: {An}
  {Application} to {StarCraft} {Micromanagement} {Tasks}.
\newblock In \emph{Proceedings of the International Conference on Learning
  Representations (ICLR)}, 2017.

\bibitem[Vinyals et~al.(2017)Vinyals, Ewalds, Bartunov, Georgiev, Vezhnevets,
  Yeo, Makhzani, K{\"u}ttler, Agapiou, Schrittwieser, Quan, Gaffney, Petersen,
  Simonyan, Schaul, van Hasselt, Silver, Lillicrap, Calderone, Keet, Brunasso,
  Lawrence, Ekermo, Repp, and Tsing]{vinyals_starcraft_2017}
Vinyals, O., Ewalds, T., Bartunov, S., Georgiev, P., Vezhnevets, A.~S., Yeo,
  M., Makhzani, A., K{\"u}ttler, H., Agapiou, J., Schrittwieser, J., Quan, J.,
  Gaffney, S., Petersen, S., Simonyan, K., Schaul, T., van Hasselt, H., Silver,
  D., Lillicrap, T., Calderone, K., Keet, P., Brunasso, A., Lawrence, D.,
  Ekermo, A., Repp, J., and Tsing, R.
\newblock {StarCraft} {II}: {A} {New} {Challenge} for {Reinforcement}
  {Learning}.
\newblock \emph{arXiv preprint arXiv:1708.04782}, 2017.

\bibitem[Watkins(1989)]{watkins_learning_1989}
Watkins, C.
\newblock \emph{Learning from delayed rewards}.
\newblock PhD thesis, University of Cambridge England, 1989.

\bibitem[Yang \& Gu(2004)Yang and Gu]{yang_multiagent_2004}
Yang, E. and Gu, D.
\newblock Multiagent reinforcement learning for multi-robot systems: {A}
  survey.
\newblock Technical report, 2004.

\end{thebibliography}
\bibliographystyle{include/icml2018}

\onecolumn
\newpage
\appendix
\section{QMIX}

\subsection{Representational Complexity}

The value function class representable with QMIX includes any value function that can be factored into a non-linear monotonic combination of the agents' individual value functions in the fully observable setting. 

This follows since the mixing network is a universal function approximator of monotonic functions \cite{Dugas_2009}, and hence can represent any value function that factors into a non-linear monotonic combination of the agent's individual value functions. Additionally, we require that the agent's individual value functions order the values of the actions appropriately. By this we mean that $Q_a$ is such that $Q_a(s_t, u^a) > Q_a(s_t, u'^a) \iff Q_{tot}(s_t, (\mathbf{u}^{-a}, u^a)) > Q_{tot}(s_t, (\mathbf{u}^{-a}, u'^a))$, i.e., they can represent a function that respects the ordering of the agent's actions in the joint-action value function. Since the agents' networks are universal function approximators \cite{pinkus1999approximation}, they can represent such a $Q_a$. Hence QMIX is able to represent any value function that factors into a non-linear monotonic combination of the agent's individual value functions. 

In a Dec-POMDP, QMIX cannot necessarily represent the value function. This is because each agent's observations are no longer the full state, and thus they might not be able to distinguish the true state given their local observations. If the agent's value function ordering is then wrong, i.e., $Q_a(\tau^a, u) > Q_a(\tau^a, u')$ when $Q_{tot}(s_t, (\mathbf{u}^{-a}, u)) < Q_{tot}(s_t, (\mathbf{u}^{-a}, u'))$, then the mixing network would be unable to correctly represent $Q_{tot}$ given the monotonicity constraints. 

QMIX expands upon the linear monotonic value functions that are representable by VDN. Table \ref{table_matrix_examples}a gives an example of a monotonic value function for the simple case of a two-agent matrix game. Note that VDN is unable to represent this simple monotonic value function.

\begin{table}[h]
    \centering
    \setlength{\extrarowheight}{3pt}
    \begin{tabular}{cc|*{2}{>{\centering\arraybackslash}p{.025\linewidth}|}}
        & \multicolumn{1}{c}{} & \multicolumn{2}{c}{Agent $2$} \\
        & \multicolumn{1}{c}{} & \multicolumn{1}{c}{$A$}  & \multicolumn{1}{c}{$B$} \\ \cline{3-4} 
        \multirow{2}{*}{\rotatebox[origin=c]{90}{Agent $1$}}  & $A$ & 0 & 1 \\ \cline{3-4}
        & $ B $ & 1 & 8  \\\cline{3-4}
        & \multicolumn{1}{c}{} & \multicolumn{2}{c}{(a)} \\
    \end{tabular}~~~~~~~
    \begin{tabular}{cc|*{2}{>{\centering\arraybackslash}p{.025\linewidth}|}}
        & \multicolumn{1}{c}{} & \multicolumn{2}{c}{Agent $2$} \\
        & \multicolumn{1}{c}{} & \multicolumn{1}{c}{$A$}  & \multicolumn{1}{c}{$B$} \\ \cline{3-4}
        \multirow{2}{*}{\rotatebox[origin=c]{90}{Agent $1$}}  & $A$ & 2 & 1 \\ \cline{3-4}
        & $ B $ & 1 & 8  \\\cline{3-4}
        & \multicolumn{1}{c}{} & \multicolumn{2}{c}{(b)} \\
    \end{tabular}
    \caption{(a) An example of a monotonic payoff matrix, (b) a non-monotonic payoff matrix.}
    \label{table_matrix_examples}
\end{table}

However, the constraint in \eqref{eq:deriv-constr} prevents QMIX from representing value functions that do not factorise in such a manner. A simple example of such a value function for a two-agent matrix game is given in Table \ref{table_matrix_examples}b. Intuitively, any value function for which an agent's best action depends on the actions of the other agents \emph{at the same time step} will not factorise appropriately, and hence cannot be represented perfectly by QMIX.  \section{Two Step Game}

\subsection{Architecture and Training}

The architecture of all agent networks is a DQN with a single hidden layer comprised of $64$ units with a ReLU nonlinearity. Each agent performs independent $\epsilon$ greedy action selection, with $\epsilon=1$. We set $\gamma=0.99$. The replay buffer consists of the last 500 episodes, from which we uniformly sample a batch of size 32 for training. The target network is updated every 100 episodes. The learning rate for RMSprop is set to $5 \times 10^{-4}$. We train for $10k$ timesteps. The size of the mixing network is $8$ units. All agent networks share parameters, thus the agent id is concatenated onto each agent's observations. We do not pass the last action taken to the agent as input. Each agent receives the full state as input.

Each state is one-hot encoded. The starting state for the first timestep is State $1$. If Agent $1$ takes Action A, it transitions to State $2$ (whose payoff matrix is all $7$s). If agent $1$ takes Action B in the first timestep, it transitions to State $3$. 

\subsection{Learned Value Functions}

The learned value functions for the different methods on the Two Step Game are shown in Tables \ref{qmix_2step_game_all} and \ref{tab:iql_qvals}. 

\begin{table}[h]
    \setlength{\extrarowheight}{3pt}{}
    \centering
    (a)
    \begin{tabular}{c|*{2}{>{\centering\arraybackslash}p{.05\linewidth}|}}
        \multicolumn{1}{c}{} & \multicolumn{2}{c}{State $1$} \\
        \multicolumn{1}{c}{} & \multicolumn{1}{c}{\bb{$A$}}  & \multicolumn{1}{c}{\bb{$B$}} \\ \cline{2-3}
        \cc{$A$} & 6.94 & 6.94 \\ \cline{2-3}
        \cc{$B$} & 6.35 & 6.36  \\\cline{2-3}
    \end{tabular}~
    \begin{tabular}{|*{2}{>{\centering\arraybackslash}p{.05\linewidth}|}}
        \multicolumn{2}{c}{State $2$A} \\
        \multicolumn{1}{c}{\bb{$A$}}  & \multicolumn{1}{c}{\bb{$B$}} \\ \cline{1-2}
        6.99 & 7.02 \\\cline{1-2}
        6.99 & 7.02  \\\cline{1-2}
    \end{tabular}~
    \begin{tabular}{|*{2}{>{\centering\arraybackslash}p{.05\linewidth}|}}
        \multicolumn{2}{c}{State $2$B} \\
        \multicolumn{1}{c}{\bb{$A$}}  & \multicolumn{1}{c}{\bb{$B$}} \\\cline{1-2}
        \text{-1.87} & 2.31 \\\cline{1-2}
        2.33 & 6.51  \\\cline{1-2}
    \end{tabular}\\\bigskip

    (b)
    \begin{tabular}{c|*{2}{>{\centering\arraybackslash}p{.05\linewidth}|}}
        \multicolumn{1}{c}{} & \multicolumn{1}{c}{\bb{$A$}}  & \multicolumn{1}{c}{\bb{$B$}} \\ \cline{2-3}
        \cc{$A$} & 6.93 & 6.93  \\ \cline{2-3}
        \cc{$B$} & 7.92 & 7.92  \\\cline{2-3}
    \end{tabular}~
    \begin{tabular}{|*{2}{>{\centering\arraybackslash}p{.05\linewidth}|}}
        \multicolumn{1}{c}{\bb{$A$}}  & \multicolumn{1}{c}{\bb{$B$}} \\ \cline{1-2}
        7.00 & 7.00 \\ \cline{1-2}
        7.00 & 7.00  \\\cline{1-2}
    \end{tabular}~
    \begin{tabular}{|*{2}{>{\centering\arraybackslash}p{.05\linewidth}|}}
        \multicolumn{1}{c}{\bb{$A$}}  & \multicolumn{1}{c}{\bb{$B$}} \\\cline{1-2}
        0.00 & 1.00 \\\cline{1-2}
        1.00 & 8.00 \\\cline{1-2}
    \end{tabular}\\\bigskip

    (c)
    \begin{tabular}{c|*{2}{>{\centering\arraybackslash}p{.05\linewidth}|}}
        \multicolumn{1}{c}{} & \multicolumn{1}{c}{\bb{$A$}}  & \multicolumn{1}{c}{\bb{$B$}} \\ \cline{2-3}
        \cc{$A$} & 6.94 & 6.93  \\ \cline{2-3}
        \cc{$B$} & 7.93 & 7.92  \\\cline{2-3}
    \end{tabular}~
    \begin{tabular}{|*{2}{>{\centering\arraybackslash}p{.05\linewidth}|}}
        \multicolumn{1}{c}{\bb{$A$}}  & \multicolumn{1}{c}{\bb{$B$}} \\ \cline{1-2}
        7.03 & 7.02 \\ \cline{1-2}
        7.02 & 7.01  \\\cline{1-2}
    \end{tabular}~
    \begin{tabular}{|*{2}{>{\centering\arraybackslash}p{.05\linewidth}|}}
        \multicolumn{1}{c}{\bb{$A$}}  & \multicolumn{1}{c}{\bb{$B$}} \\\cline{1-2}
        0.00 & 1.01 \\\cline{1-2}
        1.01 & 8.02 \\\cline{1-2}
    \end{tabular}\\\bigskip

    (d)
    \begin{tabular}{c|*{2}{>{\centering\arraybackslash}p{.05\linewidth}|}}
        \multicolumn{1}{c}{} & \multicolumn{1}{c}{\bb{$A$}}  & \multicolumn{1}{c}{\bb{$B$}} \\ \cline{2-3}
        \cc{$A$} & 6.98 & 6.97  \\ \cline{2-3}
        \cc{$B$} & 6.37 & 6.36  \\\cline{2-3}
    \end{tabular}~
    \begin{tabular}{|*{2}{>{\centering\arraybackslash}p{.05\linewidth}|}}
        \multicolumn{1}{c}{\bb{$A$}}  & \multicolumn{1}{c}{\bb{$B$}} \\ \cline{1-2}
        7.01 & 7.02 \\ \cline{1-2}
        7.02 & 7.04  \\\cline{1-2}
    \end{tabular}~
    \begin{tabular}{|*{2}{>{\centering\arraybackslash}p{.05\linewidth}|}}
        \multicolumn{1}{c}{\bb{$A$}}  & \multicolumn{1}{c}{\bb{$B$}} \\\cline{1-2}
        -1.39 & 2.57 \\\cline{1-2}
        2.67 & 6.58 \\\cline{1-2}
    \end{tabular}\\\bigskip

    (e)
    \begin{tabular}{c|*{2}{>{\centering\arraybackslash}p{.05\linewidth}|}}
        \multicolumn{1}{c}{} & \multicolumn{1}{c}{\bb{$A$}}  & \multicolumn{1}{c}{\bb{$B$}} \\ \cline{2-3}
        \cc{$A$} & 6.95 & 6.99  \\ \cline{2-3}
        \cc{$B$} & 6.18 & 6.22  \\\cline{2-3}
    \end{tabular}~
    \begin{tabular}{|*{2}{>{\centering\arraybackslash}p{.05\linewidth}|}}
        \multicolumn{1}{c}{\bb{$A$}}  & \multicolumn{1}{c}{\bb{$B$}} \\ \cline{1-2}
        6.99 & 7.06 \\ \cline{1-2}
        7.01 & 7.09  \\\cline{1-2}
    \end{tabular}~
    \begin{tabular}{|*{2}{>{\centering\arraybackslash}p{.05\linewidth}|}}
        \multicolumn{1}{c}{\bb{$A$}}  & \multicolumn{1}{c}{\bb{$B$}} \\\cline{1-2}
        -1.21 & 2.73 \\\cline{1-2}
        2.46 & 6.40 \\\cline{1-2}
    \end{tabular}\\\bigskip

    \caption{$Q_{tot}$ on the 2 step game for (a) VDN, (b) QMIX, (c) QMIX-NS, (d) QMIX-Lin and (e) VDN-S}
    \label{qmix_2step_game_all}
\end{table}

\begin{table}[h]
    \setlength{\extrarowheight}{3pt}
    \centering

    \begin{tabular}{c|*{2}{>{\centering\arraybackslash}p{.05\linewidth}|}}
        \multicolumn{1}{c}{} & \multicolumn{1}{c}{$A$}  & \multicolumn{1}{c}{$B$} \\ \cline{2-3}
        Agent $1$ & 6.96 & 4.47  \\ \cline{2-3}
    \end{tabular}~
    \begin{tabular}{|*{2}{>{\centering\arraybackslash}p{.05\linewidth}|}}
        \multicolumn{1}{c}{$A$}  & \multicolumn{1}{c}{$B$} \\ \cline{1-2}
        6.98 & 7.00 \\ \cline{1-2}
    \end{tabular}~
    \begin{tabular}{|*{2}{>{\centering\arraybackslash}p{.05\linewidth}|}}
        \multicolumn{1}{c}{$A$}  & \multicolumn{1}{c}{$B$} \\\cline{1-2}
        0.50 & 4.50 \\\cline{1-2}
    \end{tabular}\\\bigskip

    \begin{tabular}{c|*{2}{>{\centering\arraybackslash}p{.05\linewidth}|}}
        \multicolumn{1}{c}{} & \multicolumn{1}{c}{$A$}  & \multicolumn{1}{c}{$B$} \\ \cline{2-3}
        Agent $2$ & 5.70 & 5.78  \\\cline{2-3}
    \end{tabular}~
    \begin{tabular}{|*{2}{>{\centering\arraybackslash}p{.05\linewidth}|}}
        \multicolumn{1}{c}{$A$}  & \multicolumn{1}{c}{$B$} \\ \cline{1-2}
        7.00 & 7.02  \\\cline{1-2}
    \end{tabular}~
    \begin{tabular}{|*{2}{>{\centering\arraybackslash}p{.05\linewidth}|}}
        \multicolumn{1}{c}{$A$}  & \multicolumn{1}{c}{$B$} \\\cline{1-2}
        0.50 & 4.47 \\\cline{1-2}
    \end{tabular}

    \caption{$Q_{a}$ for IQL on the 2 step game}
    \label{tab:iql_qvals}
\end{table}

\subsection{Results}

\begin{figure*}[htb!]
    \centering
    \includegraphics[width=0.39\textwidth]{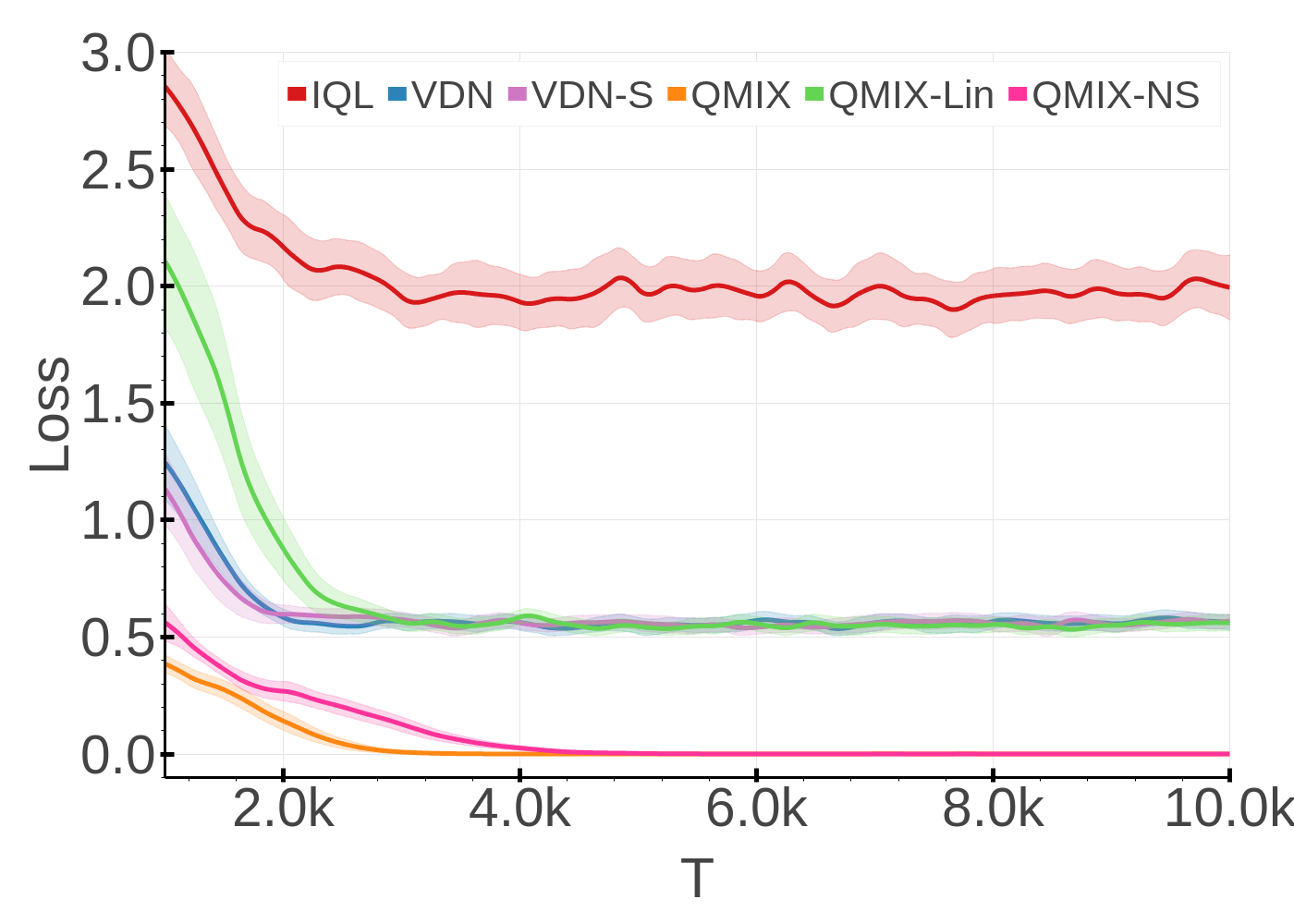}
    \caption{Loss for all six methods on the Two Step Game. The mean and 95\% confidence interval is shown across 30 independent runs.}
    \label{fig:two_step_loss}
\end{figure*}

Figure \ref{fig:two_step_loss} shows the loss for the different methods. Table \ref{table:final_test_two_step} shows the final testing reward for each method.

\begin{table}[h]
	\setlength{\extrarowheight}{3pt}
	\centering
    \begin{center}
        \begin{tabular}{|c| c | c | c | c | c |}
        \hline
        \textbf{IQL} & \textbf{VDN} & \textbf{VDN-S} & \textbf{QMIX} & \textbf{QMIX-Lin} & \textbf{QMIX-NS} \\
        \hline
        7 & 7 & 7 & 8 & 7 & 8 \\
        \hline
        \end{tabular}
    \end{center}
    \caption{The final Test Reward acheived.}
    \label{table:final_test_two_step}
\end{table}
 \section{StarCraft II Setup}

\subsection{Environment Features}
The local observations of individual agents are drawn within their field of view, which encompasses the circular area of the map surrounding units and has a radius equal to the sight range. Each agent receives as input a vector consisting of the following features for all units in its field of view (both allied and enemy): \texttt{distance}, \texttt{relative x}, \texttt{relative y} and \texttt{unit\_type}.\footnote{\texttt{unit\_type} is only included in the 2s\_3z, 3s\_5z and 1c\_3s\_5z maps.} 

The global state, which is hidden from agents, is a vector comprised of features of units from the entire map. It does not contain the absolute distances between agents and stores only the coordinates of units relative to the centre of the map. In addition, the global state includes the \texttt{health}, \texttt{shield} and \texttt{cooldown} of all units.\footnote{A unit's \texttt{cooldown} is the time it must wait before firing again. Shields act as additional forms of hit points and are lost first. In contrast to health, shields regenerate over time after absorbing damage.} In addition, the global state contains the last actions taken by all allied agents. Marines, Stalkers, Zealots, and Colossi have $45$, $80$, $100$, and $200$ hit points, respectively. In addition, Stalkers, Zealots, and Colossi have $80$, $50$, and $150$ shield points, respectively. All features, whether in local observations or global state, are normalised by their maximum values. For all unit types, the agent sight range and shooting ranges are set to 9 and 6, respectively.

\subsection{Architecture and Training}

The architecture of all agent networks is a DRQN with a recurrent layer 
comprised of a GRU with a 64-dimensional hidden state, with a fully-connected 
layer before and after.
Exploration is performed during training using independent $\epsilon$-greedy action selection, where each agent $a$ performs $\epsilon$-greedy action selection over its own $Q_a$. 
Throughout the training, we anneal $\epsilon$ linearly from $1.0$ to $0.05$ over $50k$ time steps and keep it constant for the rest of the learning. 
We set $\gamma = 0.99$ for all experiments.
The replay buffer contains the most recent $5000$ episodes.  
We sample batches of 32 episodes uniformly from the replay buffer and train on fully unrolled episodes.
The target networks are updated after every $200$ training episodes.

To speed up the learning, we share the parameters of the agent networks across all agents. 
Because of this, a one-hot encoding of the \texttt{agent\_id} is concatenated onto each agent's observations. 
All neural networks are trained using RMSprop\footnote{We set $\alpha = 0.99$ and do not use weight decay or momentum.} with learning rate $5 \times 10^{-4}$. 

During training and testing, we restrict each episode to have a length of $60$ time steps for 3m and 5m maps, $120$ time steps for 8m and 2s\_3z maps, $150$ for 3s\_5z and $200$ for 1c\_3s\_5z. If both armies are alive at the end of the episode, we count it as a loss. The episode terminates after one army has been defeated, or the time limit has been reached. 

The mixing network consists of a single hidden layer of $32$ units, utilising an ELU non-linearity. The hypernetworks are then sized to produce weights of appropriate size. The hypernetwork producing the final bias of the mixing network consists of a single hidden layer of $32$ units with a ReLU non-linearity.
 \section{StarCraft II Results}

The results for all six methods and the heuristic-based algorithm on the six maps.

\begin{table}[h]
	\setlength{\extrarowheight}{3pt}
	\centering
    \begin{center}
        \begin{tabular}{| c | c | c | c | c | c |}
        \hline
        \textbf{3m} & \textbf{5m} & \textbf{8m} & \textbf{2s\_3z} & \textbf{3s\_5z} & \textbf{1c\_3s\_5z} \\
        \hline
        76 & 60 & 95 & 82 & 45 & 70 \\ 
        \hline
        \end{tabular}
    \end{center}
    \caption{The Test Win Rate \% of the heuristic-based algorithm on the six maps.}
\end{table}

\begin{figure*}[htb!]
    \centering
    \includegraphics[height=0.4cm]{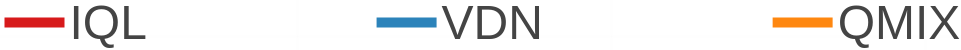}
    \vfill
    \subfigure[3m]{
        \includegraphics[width=0.315\textwidth]{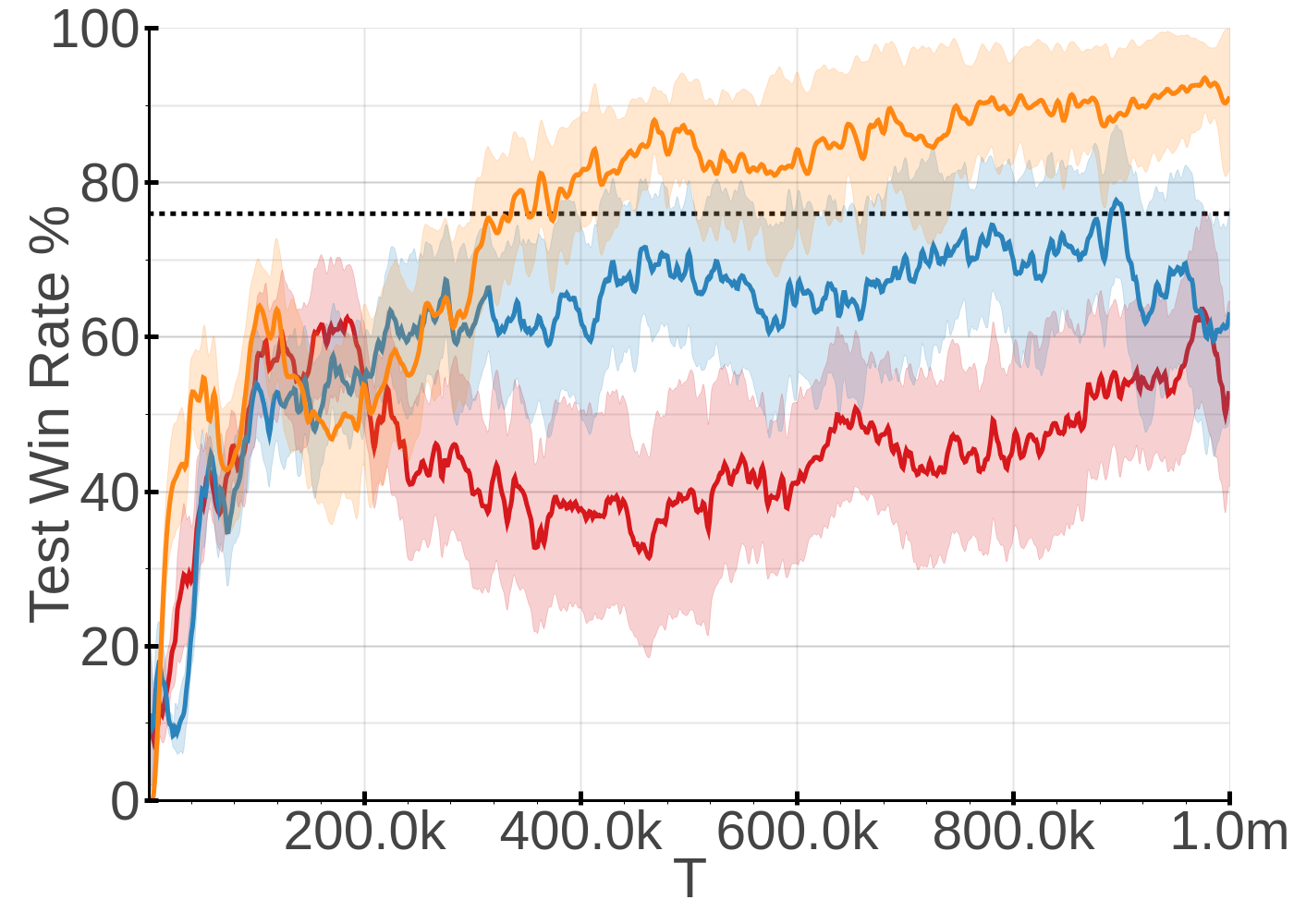}
    }
    \subfigure[5m]{
        \includegraphics[width=0.315\textwidth]{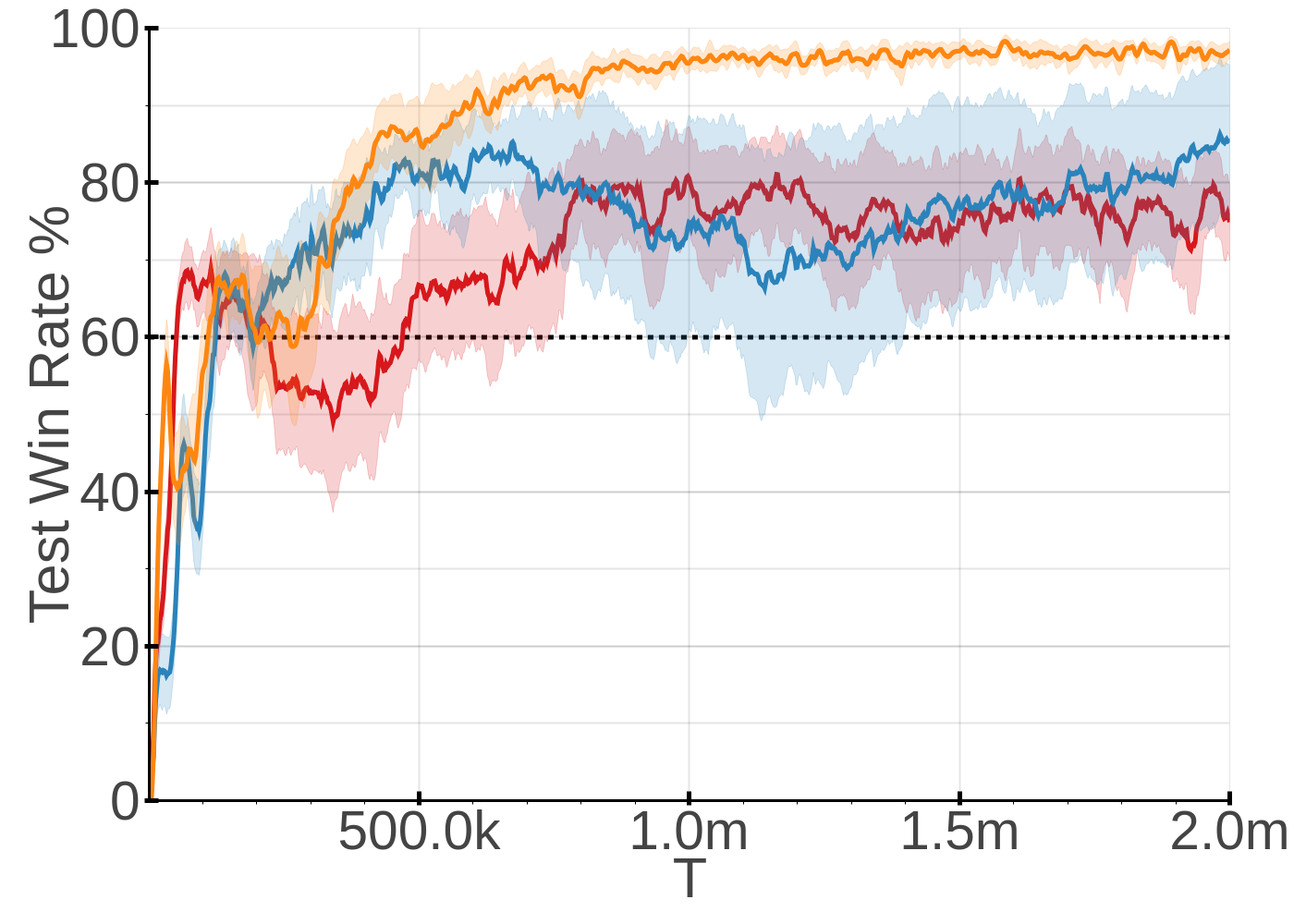}
    }
    \subfigure[8m]{
        \includegraphics[width=0.315\textwidth]{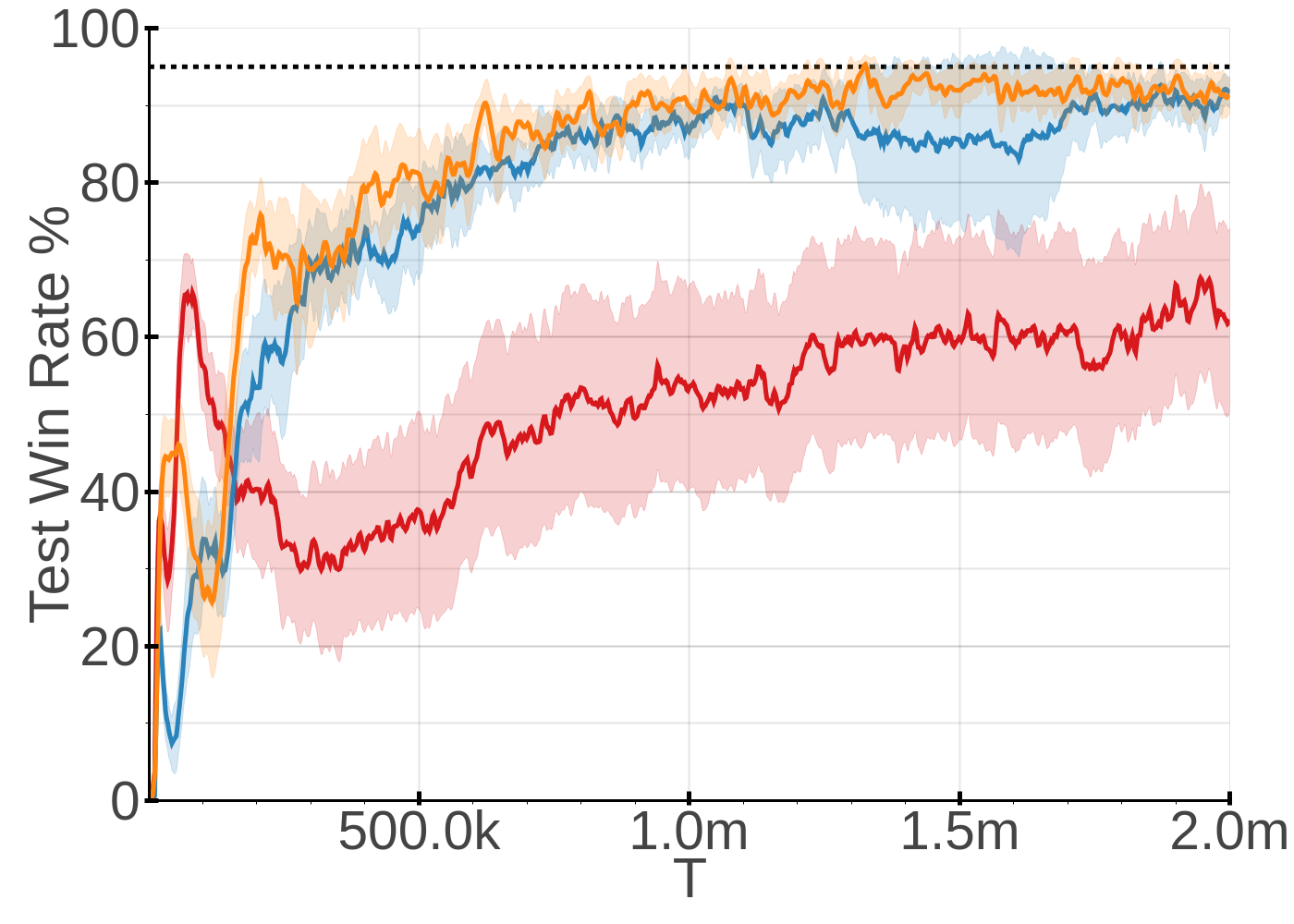}
    }
    \\
    \subfigure[2s\_3z]{
        \includegraphics[width=0.315\textwidth]{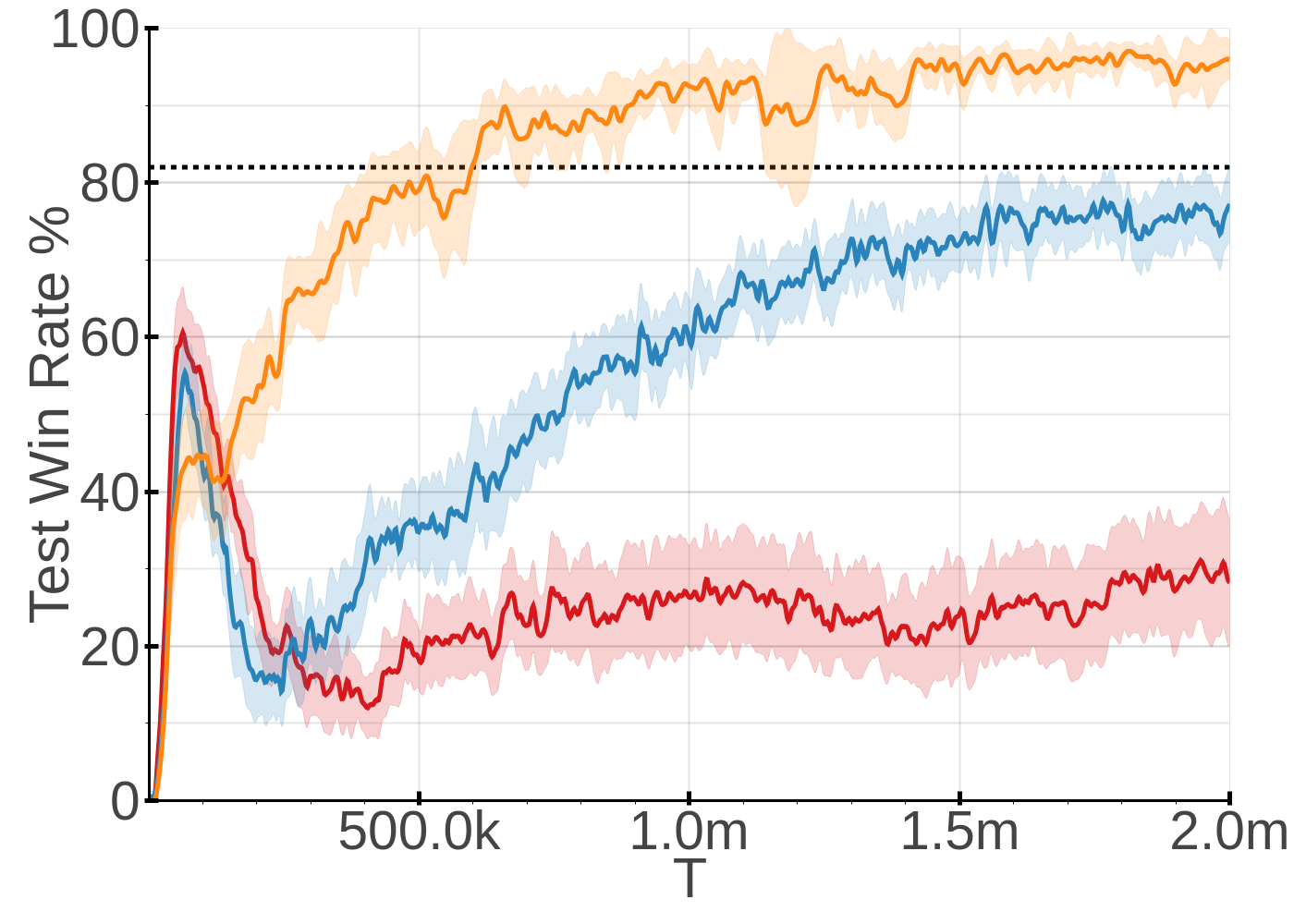}
    }
    \subfigure[3s\_5z]{
        \includegraphics[width=0.315\textwidth]{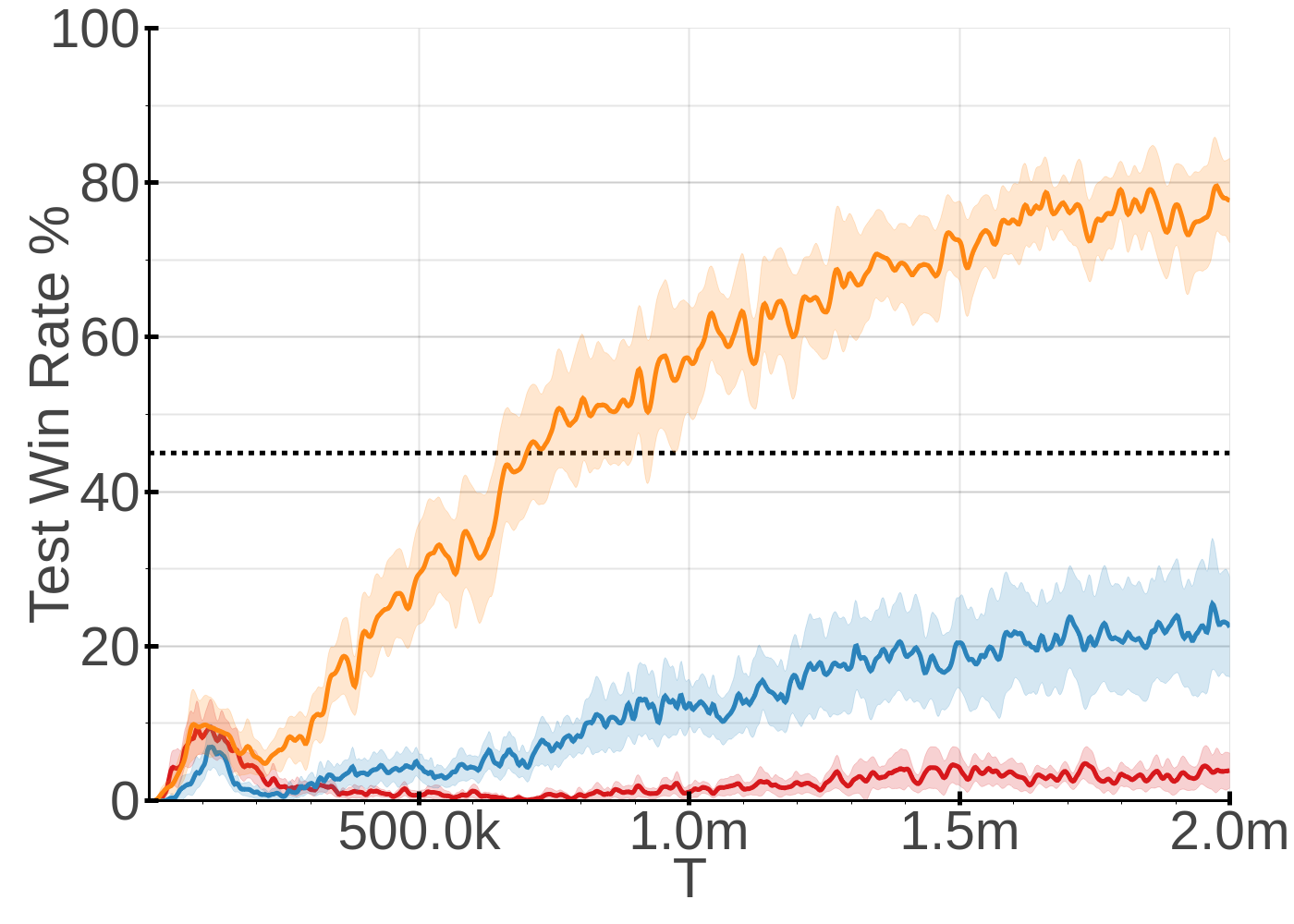}
    }
    \subfigure[1c\_3s\_5z]{
        \includegraphics[width=0.315\textwidth]{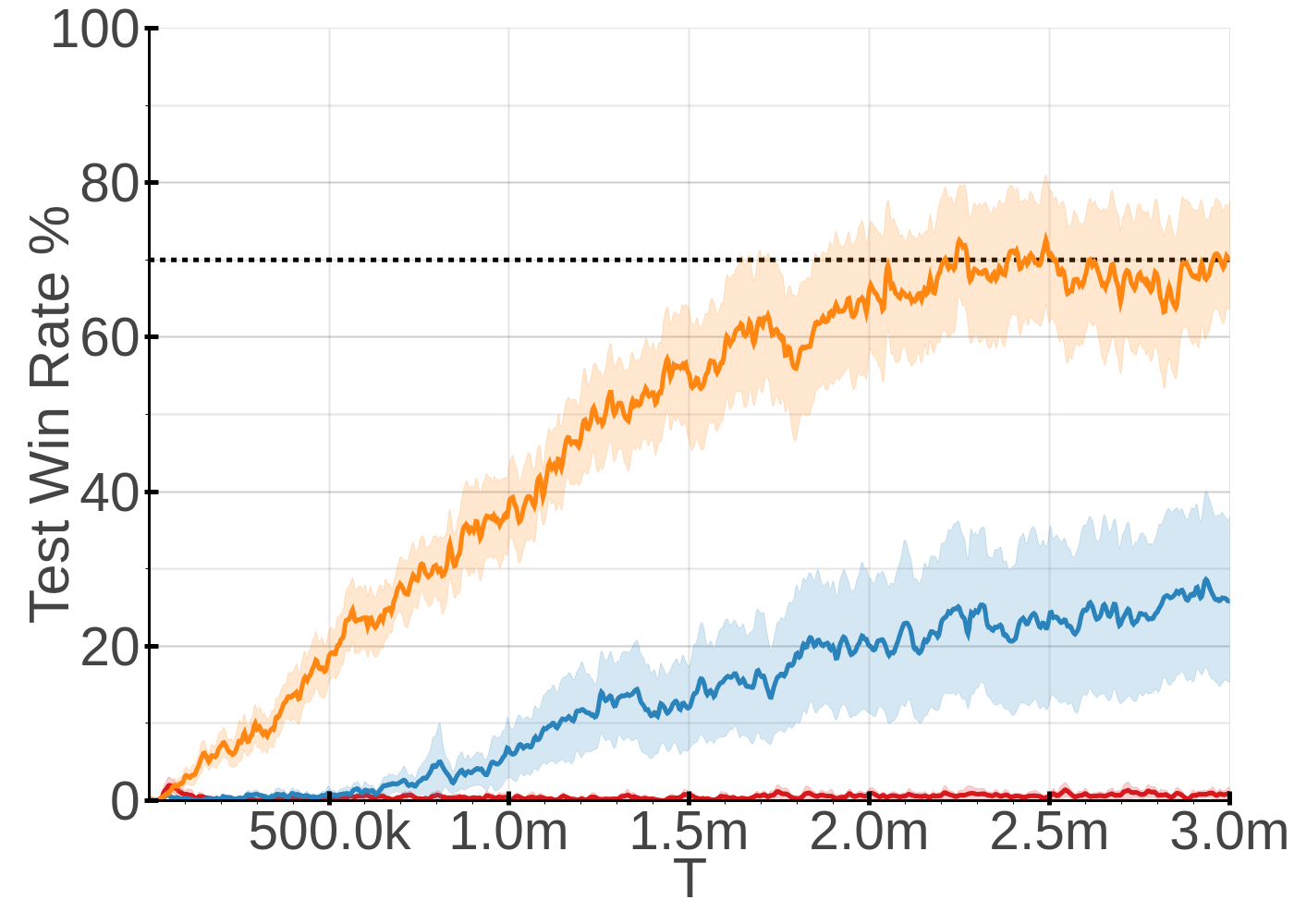}
    }
    \caption{Win rates for IQL, VDN, and QMIX on six different combat maps. The performance of the heuristic-based algorithm is shown as a dashed line.}
\end{figure*}

\begin{figure*}[htb!]
    \centering
    \includegraphics[height=0.4cm]{figures/plots/ablations_legend}
    \vfill
    \subfigure[3m]{
        \includegraphics[width=0.315\textwidth]{figures/plots/3m_3m/ablations_vdns}
    }
    \subfigure[5m]{
        \includegraphics[width=0.315\textwidth]{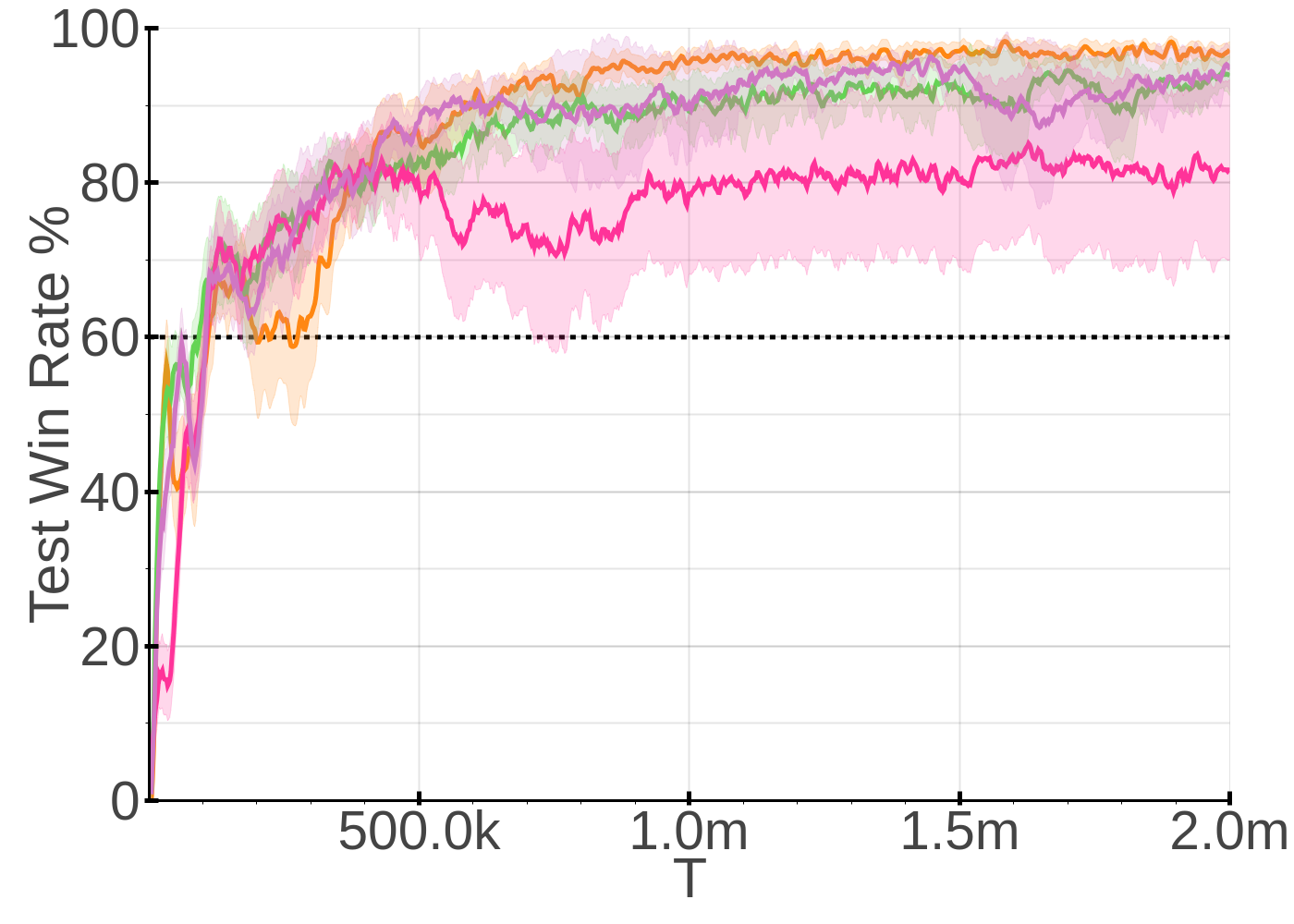}
    }
    \subfigure[8m]{
        \includegraphics[width=0.315\textwidth]{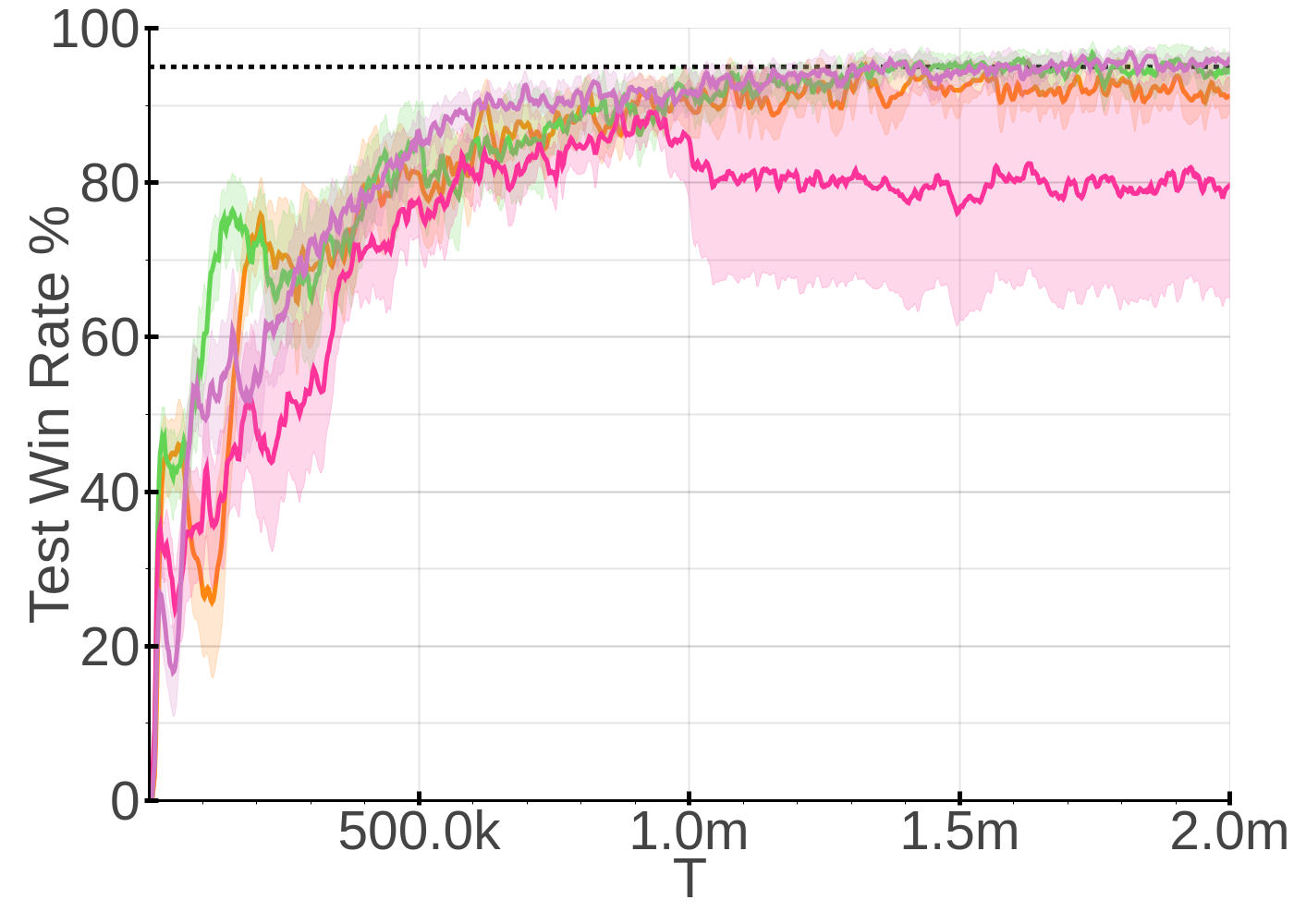}
    }
    \\
    \subfigure[2s\_3z]{
        \includegraphics[width=0.315\textwidth]{figures/plots/2d_3z/ablations_vdns}
    }
    \subfigure[3s\_5z]{
        \includegraphics[width=0.315\textwidth]{figures/plots/3d_5z/ablations_vdns}
    }
    \subfigure[1c\_3s\_5z]{
        \includegraphics[width=0.315\textwidth]{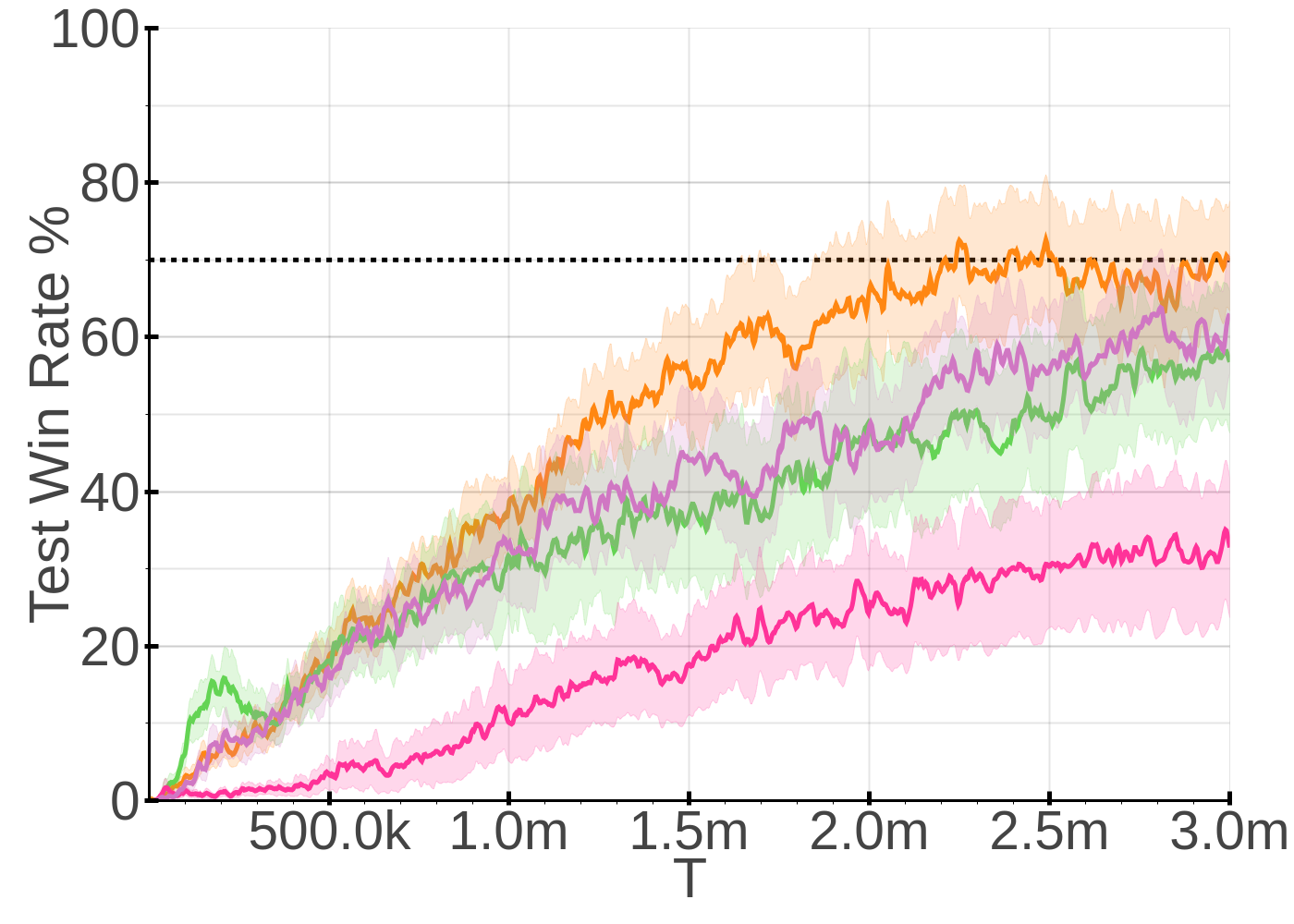}
    }
    \caption{Win rates for QMIX and ablations on six different combat maps. The performance of the heuristic-based algorithm is shown as a dashed line.}
\end{figure*}

\end{document}